\documentclass{ieeeaccess}
\usepackage{amsmath,amssymb,amsfonts}
\usepackage{algorithmic}
\usepackage{graphicx}
\usepackage{textcomp}
\def\BibTeX{{\rm B\kern-.05em{\sc i\kern-.025em b}\kern-.08em
    T\kern-.1667em\lower.7ex\hbox{E}\kern-.125emX}}
\usepackage{algorithm}
\usepackage{algorithmic}
\usepackage{subcaption}
\usepackage{multirow}
\usepackage{enumitem}
\usepackage[backend=biber]{biblatex}

\def\hypass{{HyPASS}}
\def\reid{{re-ID}}

\newcommand*{\Scale}[2][4]{\scalebox{#1}{$#2$}}%

\newcommand{\specialcell}[2][c]{%
  \begin{tabular}[#1]{@{}c@{}}#2\end{tabular}}
\newcommand{\argmin}{\mathop{\mathrm{argmin}}}
  
\begin{document}
\history{Date of publication xxxx 00, 0000, date of current version xxxx 00, 0000.}
\doi{10.1109/ACCESS.2017.DOI}

\title{Improving Unsupervised Domain Adaptive Re-Identification via Source-Guided Selection of Pseudo-Labeling Hyperparameters}
\author{\uppercase{Fabian DUBOURVIEUX \authorrefmark{1}\authorrefmark{2},
Angélique LOESCH\authorrefmark{1}, Romaric AUDIGIER \authorrefmark{1}, Samia AINOUZ\authorrefmark{2}, Stéphane CANU\authorrefmark{2}}}
\address[1]{\textit{Université Paris-Saclay, CEA, List}, \\ \textit{F-91120, Palaiseau, France}\\
    \{firstname.lastname\}@cea.fr\\}
\address[2]{\textit{Normandie Univ, INSA Rouen, LITIS}\\ 
Av. de l'Université le Madrillet 76801 Saint Etienne du Rouvray, France\\ \{firstname.lastname\}@insa-rouen.fr\\}

\markboth
{Author \headeretal: Preparation of Papers for IEEE TRANSACTIONS and JOURNALS}
{Author \headeretal: Preparation of Papers for IEEE TRANSACTIONS and JOURNALS}

\corresp{Corresponding author: Fabian DUBOURVIEUX (e-mail: fabian.dubourvieux@cea.fr).}

\begin{abstract}
Unsupervised Domain Adaptation (UDA) for re-identification (re-ID) is a challenging task: to avoid a costly annotation of additional data, it aims at transferring knowledge from a domain with annotated data to a domain of interest with only unlabeled data. Pseudo-labeling approaches have proven to be effective for UDA re-ID. However, the effectiveness of these approaches heavily depends on the choice of some hyperparameters (HP) that affect the generation of pseudo-labels by clustering. The lack of annotation in the domain of interest makes this choice non-trivial. Current approaches simply reuse the same empirical value for all adaptation tasks and regardless of the target data representation that changes through pseudo-labeling training phases. As this simplistic choice may limit their performance, we aim at addressing this issue. We propose new theoretical grounds on HP selection for clustering UDA re-ID as well as method of automatic and cyclic HP tuning for pseudo-labeling UDA clustering: HyPASS. HyPASS consists in incorporating two modules in pseudo-labeling methods: (i) HP selection based on a labeled source validation set and (ii) conditional domain alignment of feature discriminativeness to improve HP selection based on source samples.
Experiments on commonly used person re-ID and vehicle re-ID datasets show that our proposed HyPASS consistently 
improves the best state-of-the-art methods in re-ID  compared to the commonly used empirical HP setting.
\end{abstract}

\begin{keywords}
hyperparameter tuning, object re-identification, pseudo-labeling, unsupervised domain adaptation
\end{keywords}

\titlepgskip=-15pt

\maketitle

\section{Introduction}
\label{sec:intro}

\begin{figure}[t]
\includegraphics[width=1\columnwidth]{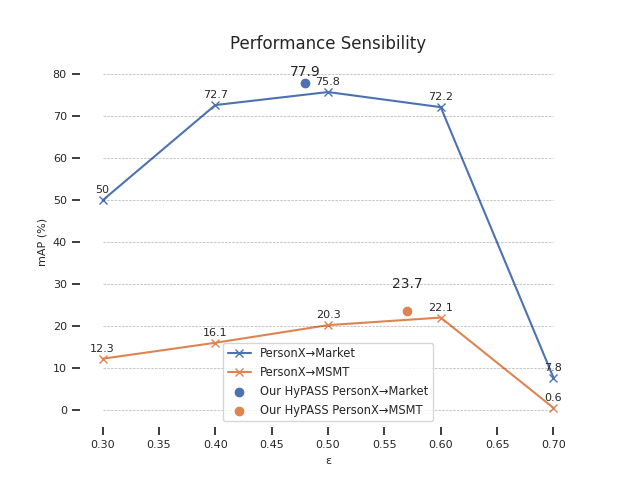}
\caption{Performance sensibility of the best state-of-the-art methods SpCL \cite{ge2020self} with respect to parameter $\epsilon$ (the maximum neighborhood distance) of DBSCAN \cite{ester1996density} for two different cross-dataset experiments. \hypass{} consistently ensures a better HP choice.}
    \label{fig:sensibility}
\end{figure}

Re-identification (\reid) aims at retrieving images of a person or an object of interest captured by different cameras. While supervised learning has achieved excellent performance on widely used \reid{} datasets \cite{ye2021deep}, it suffers from a significant drop in performance when \reid{} models are evaluated cross-dataset, i.e., on images of a target context different from the training context. To avoid a manual annotation, the computer vision community has become increasingly involved to seek how to transfer knowledge of a \reid{} model from a source domain to a target domain without identity (ID) label on the target domain.  Creative Unsupervised Domain Adaptation (UDA) methods for \reid{} have been designed. These methods tailor the open-set nature of \reid{}  in which classes of individuals at test time are different from those seen during the training stage.\\
In particular, pseudo-labeling approaches have proven to be the best UDA methods to learn ID-discriminative features for the target domain \cite{zhong19enc} \cite{song2020unsupervised} \cite{ge2020self}. For this purpose, these methods rely on generating artificial labels for the target unlabeled training data. Due to the open-set nature of the \reid{}
UDA task, pseudo-labels are generally generated by clustering the target training samples \cite{song2020unsupervised, ge2020mutual, ge2020self}. To this end, it is necessary to specify values for the hyperparameters (\textbf{HP}) that set the clustering algorithm. Density-based clustering algorithms \cite{beeferman2000agglomerative, McInnes2017, ester1996density} are the most widespread in the UDA re-ID literature. In particular, DBSCAN  \cite{ester1996density} is used for its effectiveness in a large majority of pseudo-labeling approaches, including the best performing ones \cite{ge2020self, zhai2020multiple}.
For DBSCAN, one hyperparameter to set is $\epsilon$, defined as the maximum neighborhood distance. Despite the development of approaches robust to noise in pseudo-labels \cite{ge2020mutual, ge2020self}, their final performance is still quite sensitive to the choice of $\epsilon$. In Fig.~\ref{fig:sensibility}, there is a limited range of $\epsilon$ values for which performance of SpCL \cite{ge2020self}, the best state-of-the-art methods, remain near `optimal' and not very sensitive. Indeed, given a cross-dataset task, for example PersonX$\rightarrow$Market (the re-ID datasets are presented later in Sec.~\ref{sec:dataset}), these values seem condensed in a range around $\epsilon = 0.5$, where performance reaches a mAP of $75.8 \%$. However, if $\epsilon$ is set to $0.6$, performance drops to $72.2\%$. For $\epsilon = 0.7$, the performance drop is even sharper: down to $7.8\%$.\\
\newline Therefore, selecting a suitable value for this critical HP is crucial to obtain the best performance. This behavior is not specific to DBSCAN and the same can be said for HP k of k-means (this will be discussed later in Sec.~\ref{sec:clustering} with Fig.~\ref{fig:sensibility_mmt}).
The lack of labels for the target data makes this selection non-trivial in the UDA context. Unlike the supervised setting, it is impossible to form a labeled validation set to do HP tuning with a \reid{} performance metric on the target domain (mAP, rank-1...). The state-of-the-art for UDA re-ID \cite{song2020unsupervised,ge2020self} sets these critical pseudo-labeling HP (like $\epsilon$) by validation on one adaptation task (e.g. PersonX$\rightarrow$MSMT) with a \emph{labeled} target validation data set, then uses this empirical value for other adaptation tasks. This empirical setting strategy assumes that a value selected for HP from one adaptation task transfers well to another one. However, this assumption only holds to a certain extent and, to our knowledge, there is no rule to know in advance how well this value transfers to a new task in the UDA setting. In Fig.~\ref{fig:sensibility}, by using this strategy for SpCL \cite{ge2020self} method, with the best value $\epsilon$ on PersonX$\rightarrow$MSMT ($\epsilon = 0.6$), we get a mAP of $72.2\%$ on the PersonX$\rightarrow$Market task. However, if we had chosen $\epsilon = 0.5$ we could have obtained a better mAP of $75.8\%$. This indicates that empirical setting has its limits and that a task-specific choice of HP would be more desirable in order to get maximum performance of the pseudo-labeling method. Again, these remarks also apply to other clustering algorithms (see \cite{ge2020mutual} and Fig.~\ref{fig:sensibility_mmt} for k-means).
Moreover, the clusters depend on the learned feature representation. As the feature representation varies through learning, this HP choice might even be better if we could cyclically adjust its value to the learned feature representation before each pseudo-labeling updates by clustering.\\
\newline
Motivated by the above concerns, we propose to improve existing pseudo-labeling methods by an automatic and cyclic selection of clustering HP suitable to the adaptation task and feature representation. To achieve this goal, our contribution is twofold: 
\begin{itemize}
    \item Theoretical modeling and insights that shed light on the conditions under which source-based validation is relevant for the UDA re-ID clustering task are provided.
    \item A novel method to automate the selection of clustering HP used by pseudo-labeling approaches is proposed: HyperParameters Automated by Source \& Similarities (\hypass{}). It consists in (i) a source-guided automatic HP tuning performed before each clustering phase and (ii) a conditional domain alignment of feature similarities with source ID-discriminative features applied during the training phase to improve HP selection.
\end{itemize}
Extensive experiments on commonly used and challenging \reid{} tasks for people or vehicles and ablative studies show that \hypass{} can be integrated into 
the best pseudo-labeling methods and improves consistently re-ID performance compared to a less well-chosen HP value with empirical setting.
The paper is structured as follows: In Sec.~\ref{sec:related} , we review the literature on UDA re-ID and HP selection for UDA classification. Then, in Sec.~\ref{sec:theory}, we present our theoretical grounds on clustering HP selection in the UDA re-ID setting. This motivates the design of \hypass{} presented in Sec.~\ref{sec:algo_practice}. In Sec.~\ref{sec:experiments}, \hypass{} is evaluated on commonly used and challenging cross-dataset benchmarks, and thorough analysis and discussion are conducted about its components and training computation time.
\newpage
\section{Related Work}
\label{sec:related}

\subsection{Unsupervised Domain Adaptation for \reid}
\label{sec:uda_reid}
State-of-the-art methods for UDA \reid{} can be divided into two main families: \textit{Domain translation} and \textit{Pseudo-labeling} methods.

\subsubsection{Domain Translation Methods}
\label{sec:domain_translation}
In the one hand, \textit{Image-to-Image translation} methods aim at reducing the domain discrepancy at the pixel level. A generative model \cite{zhu2017unpaired} learns to translate images from one domain to another while preserving some class-related information. Source images are translated into target style then used with their original labels to learn a \reid{} model for the target domain in a supervised way \cite{wei2018person, deng2018image, peng2019cross}. Existing works \cite{zhong2018generalizing, qi2019novel} further reduce the domain discrepancy at the camera level, with additional target camera labels. But overall, images translated into the target style highly depend on the quality of the generated images and the source domain appearances, thus failing to capture specific target \reid{} cues.
In the other hand, domain discrepancy can also be directly tackled at the feature level with \textit{Domain Invariant Feature} learning. In existing works, various assumptions are made to learn a domain-shared space, such as a semantic attribute feature space \cite{lin2018multi}, a \reid{} disentangled/factorized feature space \cite{chang2019disjoint, li2018adaptation, li2019cross} or a \reid{} feature space learned by regularizing the model with an unsupervised domain discrepancy loss to align the source and target feature distributions \cite{lin2018multi, mekhazni2020unsupervised}. As Image-to-Image translation, Domain Invariant Feature learning cannot learn target-specific discriminative features that are not shared with the source domain.
\subsubsection{Pseudo-Labeling Methods} 
\label{sec:pseudo_labeling}
Pseudo-labeling methods generally exploit a source-trained model to initialize pseudo-identity labels for target data. The pseudo-labels are generated by clustering the target data feature representations obtained by this model.  
Some works on pseudo-labeling define their own strategy to assign labels to target data based, for example, on similarity to a selected set of prototypes \cite{yu2019unsupervised, zhong19enc, luogeneralizing, wang2020unsupervised, lin2019bottom, zeng2020hierarchical}. 
Most pseudo-labeling methods are built on a self-learning iterative paradigm which alternates between (i) optimization for target \reid{} feature learning with the lastly optimized model on target images and (ii) pseudo-label prediction (pseudo-labeling) by feature clustering \cite{song2020unsupervised, zhang2019self, jin2020global, tang2019unsupervised, zhai2020ad, zou2020joint, yang2019asymmetric, chendeep, ge2020mutual, zhai2020multiple, zhao2020unsupervised, zou2020joint,peng2020unsupervised, Zhang_2021_CVPR, Yang_2021_CVPR}. Most of these works improve the classical self-learning algorithm on not overfitting the pseudo-label errors, by using teacher-student or ensemble of expert models \cite{ge2020mutual, zhao2020unsupervised, zhai2020multiple} while other approaches focus on designing efficient sample selection and outlier detection strategies \cite{yang2019asymmetric, chendeep}. More robust frameworks are also designed by optimizing losses based on distance distributions \cite{jin2020global, liu2020domain}, by leveraging local features \cite{fu2019self}, intra-inter camera features \cite{Xuan_2021_CVPR,lin2020unsupervised}, the labeled source samples \cite{dub2020}, multiple cluster views \cite{feng2021complementary} or attention-based model \cite{jiattention}, or by mixing pseudo-labels with domain-translation methods \cite{zhai2020ad, tang2019unsupervised, zou2020joint, Chen_2021_CVPR}, online pseudo-label refinery strategy, temporal ensembling and label propagation \cite{Zhang_2021_CVPR, Zheng_2021_CVPR} or meta learning \cite{Yang_2021_CVPR}. A recent approach, SpCL \cite{ge2020self}, proposed self-contrastive learning during the training phase, by leveraging the source and target samples.
Most of the above-mentioned pseudo-labeling methods, including the best and most recent ones, use DBSCAN to pseudo-label the target training samples \cite{song2020unsupervised, zhang2019self, jin2020global, tang2019unsupervised, zhai2020ad, zou2020joint, mekhazni2020unsupervised, yang2019asymmetric, chendeep, zhao2020unsupervised, zou2020joint, ge2020mutual, zhai2020multiple}. They are all possibly affected by the clustering sensibility to hyperparameters, as it is shown in \cite{song2020unsupervised} and illustrated in Fig. \ref{fig:sensibility}, where performance of the best state-of-the-art methods, SpCL, depends on the choice of a critical HP. Other approaches, using less common clustering algorithms, also seem concerned (shown later in Sec.~\ref{sec:clustering} with Fig.~\ref{fig:sensibility_mmt} for k-means). Moreover, to our knowledge, they all choose a fixed empirical value to set this HP, which remains the same no matter the adaptation task, and through all the pseudo-labeling cycles. performance of these approaches may suffer from this restricted HP setting. Our contribution aims at overcoming those limiting aspects by integrating a new automatic and cyclic HP selection phase into the pseudo-labeling cycle. Our contribution aims to be general so that it can be easily integrated and improve any existing or future pseudo-labeling approach.

\subsection{Hyperparameter Selection for UDA classification}
\label{sec:hp_classif}

As HP selection in the UDA setting has been studied, to our knowledge, only for the classification task, we focus on the related work for this task. In UDA classification, HP selection remains a major problem. Many approaches in UDA classification use the same strategy as UDA re-ID pseudo-labeling methods: the empirical setting of HP values, used on different cross-dataset adaptation tasks \cite{tzeng2017adversarial, pinheiro2018unsupervised, saito2018maximum, pan2020unsupervised}. Manually labeling a part of the target dataset to make a validation set \cite{hoffman2018cycada} is out of the UDA context. 
The use of a source validation set \cite{ganin2016domain, peng2018visda} offers biased estimation of the classification target expected risk because of the domain discrepancy. Importance weighting methods \cite{sugiyama2007covariate, long2018conditional, cortes2010learning} tackle this issue by weighting the estimated risk with source samples but they still suffer from high variance estimation. The recent work  \cite{you2019towards} improves these approaches and proposes an importance-weighted cross-validation in the feature space to reduce the source estimator variance. 
However, two major aspects prevent its application for HP selection of the pseudo-labeling UDA clustering. First, it requires the estimation of probability densities of the source and target distributions (in the feature space). cyclically integrated in a pseudo-labeling framework, these densities should be re-estimated before each update of the pseudo-labels by clustering. This would be harder to integrate in any pseudo-labeling methods, computationally expensive and the ratio of estimated densities could increase approximation errors. Finally, the approach is adapted for classification problems only, which differs from the clustering task. \\
To our knowledge, there is no general work on clustering HP selection adapted to UDA pseudo-labeling. That's why we recast the theory behind these source leveraging approaches \cite{sugiyama2007covariate, long2018conditional, cortes2010learning, you2019towards} to fit the clustering task. Moreover, in order to better integrate it into pseudo-labeling approaches, our approach takes a new turn compared to those ones, by avoiding estimation of importance weights: we propose to optimize the model for domain alignment in the feature similarity space with source ID-discriminative features to improve the estimation with a source validation set by reducing its variance.

\section{Theoretical Grounds of Hyperparameter Selection for Clustering in UDA \reid}
\label{sec:theory}

The selection of HP $\lambda \in \mathbb{R}^{n_{\lambda}}, n_{\lambda} \in \mathbb{N^*}$ consists in finding the value $\lambda^* \in \mathbb{R}^{n_{\lambda}}$ that minimizes a defined expected risk. Unlike the models learnable parameters, HPs are not directly learnt during the training loop of a machine learning pipeline. A typical strategy to estimate $\lambda^*$ is model selection: among a set of candidate models defined by different HP values, we choose the one that gives the lowest empirical risk. This strategy is not applicable with the UDA setting because target annotations are not available. Moreover, as discussed in Sec.~\ref{sec:hp_classif}, existing approaches (for classification) are not directly adapted for \reid{}. The goal of this section is thus to give theoretical leads that will give us more insights about two questions: How do the source data bias the target risk estimation? How to overcome this bias? We first introduce notations and the problem formulation (Sec.~\ref{sec:notations}). Then we define the expected risk to optimize for the clustering task (Sec.~\ref{sec:risk_minimization}), in order to deduce an empirical estimate based on the source data (Sec.~\ref{sec:domain_discrepancy}). Finally, a focus is given to the variance of this estimate to better understand how to improve HP selection by reducing it (Sec.~\ref{sec:variance}). For this, we first show that the variance can be reduced by reducing the domain discrepancy between the source and target in the feature similarity space (Sec.~\ref{sec:feature_sim}). Then we give theoretical analysis on the pairwise ratio, showing that with reasonable assumptions, the source empirical risk can be used directly to do efficient HP selection (Sec.~\ref{sec:weight_ratio}).

\subsection{Problem Formulation and Notations}
\label{sec:notations}
\subsubsection{Offline vs Online Cyclic HP tuning for clustering}
\label{sec:cyclic}
If we focus on the iterative pseudo-labeling paradigm, we can note that the learned feature representation changes during each training phase of an iterative cycle. Since the pseudo-labels are updated by clustering in this representation space, we intuitively expect the optimal clustering hyperparameter value to change when this representation changes (as it will be shown empirically in Sec.~\ref{sec:cluster_quality}). The model selection is classically done via an evaluation criterion on the "main" task (in our case the re-ID as a retrieval task). Proceeding in this way necessarily implies training completely with selected HP values, evaluating (with \reid{} metrics such as mAP) and repeating again and again. This would thus make the selection computationally expansive (a training time analysis is given in Sec.~\ref{sec:time}). To overcome this, our idea is to perform an online model selection directly at the clustering task level, at each iterative cycle.\\
\subsubsection{Modeling the clustering task} 
As introduced in Sec.~\ref{sec:theory}, the first step is to define the expected risk to be minimized w.r.t $\lambda$ for HP selection. This expected risk $\mathcal{R}_{\mathcal{L},p}$  (defined in \cite{vapnik1998statistical}) is defined in relation to the unknown distribution of data characterized by the probability density $p$ and a cost function $\mathcal{L}$ which depends on our underlying task: a clustering task for our problem. A clustering is considered "good" when it generates pseudo-labels related to the ground-truth identity labels. Our idea is therefore to model this clustering task as a verification problem. For this, let's suppose that the \reid{} data are i.i.d and come from an unknown joint distribution given by the density $p(x, x', r)$ defined on $\chi \times \chi \times \{-1,1\}$ where $\chi \subseteq \mathbb{R}^{n_\chi}, n_{\chi} \in \mathbb{N}$  represents the set of images for which $r=1$ if $x$ and $x'$ have the same ID and $r=-1$ otherwise. 
Thus, the goal is to find a clustering function $C_{\lambda}$ which is expected to classify all the $m \in \mathbb{N}$ pairs of images in a set $X = ((x_i,x_i')_{1 \leq i \leq m})$ as their respective ground truth labels are $R = (r_i)_{1 \leq i \leq m}$. 
We also assume that clusters are predicted from a measure of similarities between elements in the set. For the set $X$, the pairwise similarities are given by $S(X) = (s(x_i,x'_i))_{1 \leq i \leq m}$, where $s : \chi \times \chi \rightarrow \mathbb{R}$ is a given similarity function. Therefore, $C_{\lambda}$ is a $\mathbb{R}^{m} \rightarrow \{-1,1\}^{m}$ function.

\subsection{Similarity-Based Clustering Risk Minimization}
\label{sec:risk_minimization}
By definition, following previous notations, the expected risk $\mathcal{R}_{\mathcal{L},p}$ for the clustering task can be seen as a function of $\lambda$:
\begin{equation}
\mathcal{R}_{\mathcal{L},p}(\lambda) \triangleq \int_{X,R}{\mathcal{L}\bigl(C_{\lambda}(S(X)),R\bigr)p(X,R)dXdR} \; ,
\label{eq:expected}
\end{equation}

where $p(X,R)$ is a joint probability density defined on $(\chi \times \chi)^m \times \{-1,1\}^{m}$.\\
The UDA setting for the clustering task does not involve only one distribution associated to its density $p$, but two distributions related to the source $\mathcal{S}$ and the target $\mathcal{T}$. Their joint probability densities are noted respectively $p^\mathcal{S}(X,R)$ and $p^\mathcal{T}(X,R)$. To perform source-based HP selection, we need to link the target expected risk $ \mathcal{R}_{\mathcal{L},p^\mathcal{T}}$ defined by Eq.~\ref{eq:expected} with $p^\mathcal{S}$.

\subsection{Similarity Importance-Weighted Risk}
\label{sec:domain_discrepancy}
We consider the \reid{ }UDA context with the target and source distributions defined above. 
Our goal is to link the target expected risk (Eq.~\ref{eq:expected}) with $p^\mathcal{S}$. 
By developing the target expected risk, we have:
\begin{equation}
\Scale[0.9]{
\begin{aligned}
\mathcal{R}_{\mathcal{L},p^\mathcal{T}}(\lambda) &= \int_{X,R}{\mathcal{L}(C_{\lambda}(S(X)),R)p^\mathcal{T}(X,R)}dXdR\\
 &= \int_{X,R}{\frac{p^\mathcal{T}(X,R)}{p^\mathcal{S}(X,R)}\mathcal{L}(C_{\lambda}(S(X)),R)p^\mathcal{S}(X,R)dXdR}\\
 &= \int_{X,R}{w(X,R)\mathcal{L}(C_{\lambda}(S(X)),R)p^\mathcal{S}(X,R)}dXdR \\.
\end{aligned}
}
\end{equation}
With the pairwise weight ratio $w$ defined as: 
\begin{equation}
w(X,R) \triangleq \frac{p^\mathcal{T}(X,R)}{p^\mathcal{S}(X,R)}.
\end{equation}
Then we can define the pairwise weighted risk as:
\begin{equation}
\Scale[0.95]{
\begin{aligned}
 \mathcal{R}_{\mathcal{L},w}(\lambda)  {}  & \triangleq \int_{X,R}{w(X,R)\mathcal{L}(C_{\lambda}(S(X)),R)p^\mathcal{S}(X,R)}dXdR. \\
\end{aligned}
}
 \label{eq:expected_w}
\end{equation}
From Eq.~\ref{eq:expected_w}, we can deduce the associated pairwise weighted empirical risk which is an unbiased estimator of $\mathcal{R}_{\mathcal{L},p^\mathcal{T}}(\lambda)$ with finite source samples:
\begin{equation}
\mathcal{\hat{R}}_{\mathcal{L},w}(\lambda) = \frac{1}{N} \sum_{i=1}^{N}{w(X_i,R_i)\mathcal{L}(C_{\lambda}(S(X_i)),R_i)} ,
\label{eq:empirical_w}
\end{equation} where $\{(X_i,R_i)\}_{1 \leq i \leq N}, N \in \mathbb{N^*}$ are samples from $p^\mathcal{S}(X,R)$. 

\subsection{Variance of the estimator}
\label{sec:variance}

Even if the estimator given by Eq.~\ref{eq:empirical_w} is unbiased, a high variance can add noise to HP selection with source samples. Before giving an expression of the estimator's variance, we define the exponential in base 2 of the R\'{e}nyi divergence (called R\'{e}nyi divergence in the rest of the paper for simplicity) of order $\alpha \geq 0$, $\alpha \neq 1$ between the source and target distribution described by the densities $p^\mathcal{S}$ and $p^\mathcal{T}$ as:

\begin{equation}
\label{eq:renyi}
\begin{aligned}
d_{\alpha}(p^\mathcal{T}||p^\mathcal{S}) &\triangleq \left( \int_{X,R}{\frac{p^\mathcal{S}(X,R)^{\alpha}}{p^\mathcal{T}(X,R)^{\alpha-1}}}dXdR \right) ^{\frac{1}{\alpha-1}}\\
&= \left( \int_{X,R}{w(X,R)^{-\alpha}p^\mathcal{T}(X,R)dXdR} \right) ^{\frac{1}{\alpha-1}}\\
&= \left( \mathop{\mathbb{E}}_{(X,R) \sim p^\mathcal{T}}[ w(X,R)^{-\alpha} ]\right) ^{\frac{1}{\alpha-1}}.
\end{aligned}
\end{equation}

Let $Y$ be $Y=w(X,R)\mathcal{L}(C_{\lambda}(S(X)),R)$ for $(X,R) \sim p^\mathcal{S}(X,R)$. Using the lemma 2 from Cortes {\it et al.} \cite{cortes2010learning} and with the definition of $\hat{\mathcal{R}}_{\mathcal{L},w}$ (Eq.~\ref{eq:empirical_w}), we can get a bound on the variance of $\mathcal{\hat{R}}_{\mathcal{L},w}(\lambda)$:
\begin{equation}
\label{eq:variance}
\Scale[0.9]{
\begin{aligned}
Var(Y) &= \mathop{\mathbb{E}}_{(X,R) \sim p^\mathcal{S}}[ Y^2 ]-\mathop{\mathbb{E}}_{(X,R) \sim p^\mathcal{S}}[ Y ]^2\\
&\leq d_{\alpha+1}(p^\mathcal{T}||p^\mathcal{S})\mathcal{R}_{\mathcal{L},p^\mathcal{T}}(\lambda)^{1-\frac{1}{\alpha}} - \mathop{\mathbb{E}}_{(X,R) \sim p^\mathcal{S}}[ Y ]^2 \\
&\leq d_{\alpha+1}(p^\mathcal{T}||p^\mathcal{S})\mathcal{R}_{\mathcal{L},p^\mathcal{T}}(\lambda)^{1-\frac{1}{\alpha}}-
\mathcal{R}_{\mathcal{L},p^\mathcal{T}}(\lambda)^2\\
Var(\hat{\mathcal{R}}_{\mathcal{L},w}) &\leq \frac{d_{\alpha+1}(p^\mathcal{T}||p^\mathcal{S})\mathcal{R}_{\mathcal{L},p^\mathcal{T}}(\lambda)^{1-\frac{1}{\alpha}}-
\mathcal{R}_{\mathcal{L},p^\mathcal{T}}(\lambda)^2}{N} \; ,
\end{aligned}
}
\end{equation}

This bound on the empirical risk variance confirms the intuition that the more source (validation) samples we have, the lesser is the variance. However, in practice, the amount of labeled source samples is limited. Therefore we cannot act on this constant in order to improve our estimation.
However, this bound on the empirical risk variance also shows that the greater the $d_{\alpha+1}(p^\mathcal{T}||p^\mathcal{S})$, the greater the variance of the estimator. In order to control this variance, and therefore improve the use of the pairwise weighted empirical risk estimator for model selection, it is necessary to control $d_{\alpha+1}(p^\mathcal{T}||p^\mathcal{S})$ which measures the domain discrepancy between $p^\mathcal{T}$ and $p^\mathcal{S}$ according to the R\'{e}nyi divergence. Moreover, reducing this divergence should make the estimation less sensible to the number of source validation samples according to Eq.~\ref{eq:variance}.
\subsection{Addressing the variance and weight ratio}
\begin{figure*}[t!]
\centering
\includegraphics[width=2\columnwidth]{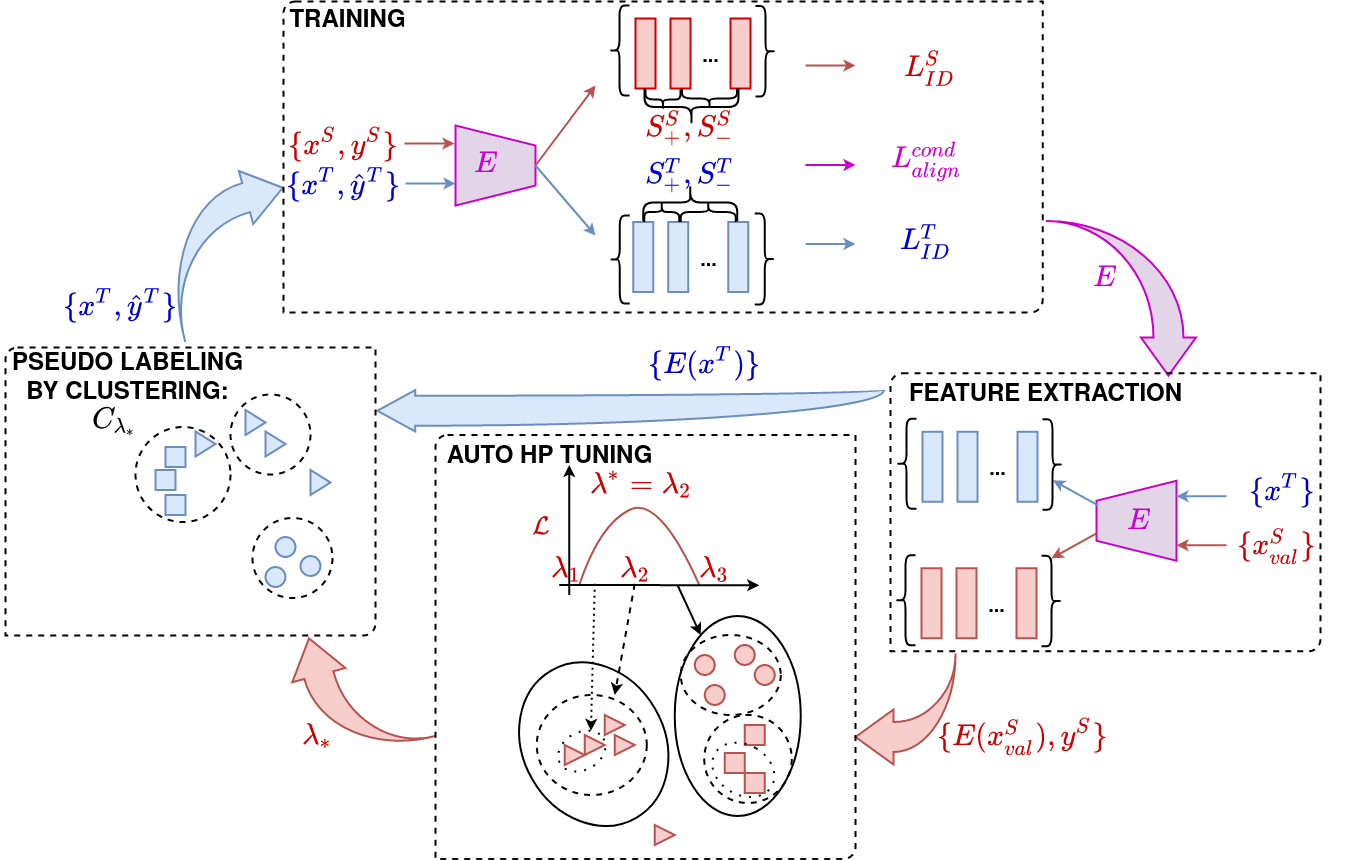}
\caption{Our HyperParameters Automated by Source \& Similarities (\hypass{}) cyclically integrated in iterations of a classical pseudo-labeling framework.}
\label{fig:framework}
\end{figure*}
\subsubsection{Using feature similarity}
\label{sec:feature_sim}
The input space (images) is high-dimensional. Therefore, $d_{\alpha+1}(p^\mathcal{T}||p^\mathcal{S})$ is more likely to be greater (and thus the variance of the estimator given by Eq.~\ref{eq:variance}) than the divergence between probability distributions in a lower-dimensional feature space ( as stated in Sec. 4.2 of \cite{you2019towards}). Indeed, the pairwise weight ratio can more likely grow to infinity since $p^\mathcal{S}$ when $p^\mathcal{T} \neq 0$ is more likely to be 0. Moreover, a feature space induced by a learnable feature encoder could allow us to reduce the divergence by penalizing it during the learning phase.\\
Usually in \reid{}, a feature space is learned so that a given similarity function used in this space can measure ID relatedness between images. Therefore, we introduce a feature encoder $E : \chi  \rightarrow \mathbb{R}^{n_E}, n_E \in \mathbb{N}$ and redefine $s : \mathbb{R}^{n_E} \times \mathbb{R}^{n_E} \rightarrow \mathbb{R}$. We also define $S_E$ the feature similarity function with respect to E such as $S_E(X) =(s(E(x_i),E(x'_i)))_{1 \leq i \leq m}$. Thus, $S_E$, projects the set of images $X$ into a new set $S \in \mathbb{R}^{m}$, in a space we call the feature similarity space. Let $p^\mathcal{S}_{S_E}(S,R)$ (resp. $p^\mathcal{T}_{S_E}(S,R)$) be the feature similarity distribution densities of $\mathcal{S}$ (resp. $\mathcal{T}$) induced by $S_E$ and defined on $\mathbb{R}^{m} \times \{-1,1\}^m$. We consider this space as our new input space for computing the risks and therefore if we note 
\begin{equation}
    w_{S_E}(S,R) = \frac{p^\mathcal{T}_{S_E}(S,R)}{p^\mathcal{S}_{S_E}(S,R)},
\end{equation}
with analogous definitions and notations, we deduce the pairwise similarity weighted empirical risk $\mathcal{\hat{R}}_{\mathcal{L},w_{S_E}}$:

\begin{equation}
\mathcal{\hat{R}}_{\mathcal{L},w_{S_E}}(\lambda) = \frac{1}{N} \sum_{i=1}^{N}{w_{S_E}(S_i,R_i)\mathcal{L}(C_{\lambda}(S_i),R_i)} ,
\label{eq:empirical_w_s}
\end{equation} where $\{(S_i,R_i)\}_{1 \leq i \leq N}, N \in \mathbb{N}$ are samples from $p^\mathcal{S}_{S_E}$.\\
In practice, we have directly access to sets of pairwise image samples $\{(X_i,R_i)\}_{1 \leq i \leq N}$ defined above and we use $S_E$ to get $\{(S_i,R_i)\} = \{(S_E(X_i),R_i)\}$.\\
According to Eq.~\ref{eq:expected_w}, $\mathcal{\hat{R}}_{\mathcal{L},w_{S_E}}$ is an unbiased estimator of the expected target risk $\mathcal{R}_{\mathcal{L},p^\mathcal{T}_{S_E}}$, that we can use to do HP selection of $\lambda$ with source labeled samples. We expect this new estimator to be better for HP selection.
Indeed, we expect it to have a lower variance than due to the lower domain discrepancy in this learnable low-dimensional feature space (as stated in Sec. 4.2 of \cite{you2019towards}):
\begin{equation}
    Var(\mathcal{\hat{R}}_{\mathcal{L},w_{S_E}}) \leq Var(\hat{\mathcal{R}}_{\mathcal{L},w}).
\end{equation}
\newline
\newline

In addition, the pairwise data samples being i.i.d. (see Sec.~\ref{sec:notations}), the pairwise similarities are i.i.d. too and therefore the densities in the feature similarity space can be written as:
\begin{center}
\label{eq:cond_densities}
$$
\left\{
    \begin{array}{ll}
       p^\mathcal{S}_{S_E}(S,R) =  p^\mathcal{S}(R) \Pi_{i=1}^{m} p^\mathcal{S}_{S_E}(S_i|R_i)
       \vspace{5pt}
       \\
       p^\mathcal{T}_{S_E}(S,R) =  p^\mathcal{T}(R) \Pi_{i=1}^{m} p^\mathcal{T}_{S_E}(S_i|R_i) \; .
    \end{array}
\right.
$$
\end{center}

Since $ p^\mathcal{S}(R)$ and $ p^\mathcal{T}(R)$ are fixed by the domain distributions and are independent from $E$, we assume that $E$ can be learned to penalize the conditional domain discrepancy (i.e. the the divergence between the conditional distributions) in the feature similarity space in order to improve HP selection with our estimator $\mathcal{\hat{R}}_{\mathcal{L},w_{S_E}}$.

\subsubsection{Computing the pairwise weight ratio}
\label{sec:weight_ratio}
To sum up, our goal is to do HP selection of $\lambda$ by minimizing $\mathcal{R}_{\mathcal{L},p^\mathcal{T}}$ (Eq.~\ref{eq:expected_w}) w.r.t $\lambda$. For this, we established the expression of the pairwise weighted empirical risk estimator $\mathcal{\hat{R}}_{\mathcal{L},w_{S_E}}$ with source samples (Eq.~\ref{eq:empirical_w_s}). This estimator will be improved by learning $E$ to penalize the conditional domain discrepancy in the feature similarity space. Using $\mathcal{\hat{R}}_{\mathcal{L},w_{S_E}}$ requires to compute $w_{S_E}$.
As mentioned in Sec.~\ref{sec:hp_classif}, unlike importance weighted risk estimation approaches for UDA classification, we do not wish to estimate the pairwise weight ratio in a pseudo-labeling framework: this would require estimating the probability density of this ratio at each new pseudo-labeling step. This would be computationally expensive. Moreover, the quotient of estimated probabilities in the ratio could increase approximation errors and therefore add noise to the risk estimate. 
To avoid computing pairwise weight ratio, it would be desirable that we can do HP selection using the source empirical risk $\mathcal{\hat{R}}_{\mathcal{L},p^\mathcal{S}_{S_E}}(\lambda)$.\\
To do relevant HP selection using $\mathcal{\hat{R}}_{\mathcal{L},p^\mathcal{S}_{S_E}}(\lambda)$ instead of $\mathcal{\hat{R}}_{\mathcal{L},w_{S_E}}(\lambda)$ , it is therefore necessary that $\argmin_{\lambda} \mathcal{\hat{R}}_{\mathcal{L},p^\mathcal{S}_{S_E}}(\lambda) \approx \argmin_{\lambda} \mathcal{\hat{R}}_{\mathcal{L},w_{S_E}}(\lambda)$.
In other words, this ensures that selecting the best $\lambda$ with $\mathcal{\hat{R}}_{\mathcal{L},p^\mathcal{S}_{S_E}}(\lambda)$ is the same as selecting the best $\lambda$ with $\mathcal{\hat{R}}_{\mathcal{L},w_{S_E}}(\lambda)$.\\ Given the expression of $\mathcal{\hat{R}}_{\mathcal{L},w_{S_E}}(\lambda)$ (Eq.~\ref{eq:empirical_w_s}), a direct sufficient condition to ensure this is that:
\begin{equation}
\label{eq:condition}
\forall 1 \leq i \leq N, w_{S_E}(S_i,R_i) = c, c \in \mathbb{R+}.\\
\end{equation}

In practice, Eq.~\ref{eq:condition} can be satisfied by using the whole source validation set as a unique pair $(S,R)$ to do HP selection. This will be part of our framework design choice as discussed later in Sec.~\ref{sec:ari} in what we call One-clustering evaluation.

To summarize, these theoretical considerations show us that to select HP $\lambda$ from the source examples, it is sufficient to minimize the source empirical risk, provided that we satisfy condition (i) and that we minimize the conditional domain discrepancy in the feature similarity space w.r.t $E$.
\section{Source-Guided Selection of Pseudo-Labeling Hyperparameters and Similarity Alignment}
\label{sec:algo_practice}
We wish to apply the theory discussed above and integrate it into a pseudo-labeling algorithm. For this purpose, we propose a novel method integrated intto the classical iterative pseudo-labeling paradigm \cite{song2020unsupervised}: HyperParameters Automated by Source \& Similarities (HyPASS).
Fig.~\ref{fig:framework} gives an overview of the incremented method.
HyPASS consists in integrating a new clustering HP selection phase (Auto HP TUNING) from a source validation set before each clustering update and optimizing the model to minimize the conditional feature similarity domain discrepancy $L^{cond}_{align}$. In this part, we give more details about this two major novelties.

\subsection{Automatic Clustering HP Tuning}
\label{sec:ari}

Our method proposes a new step of automatic selection of clustering HP $\lambda$. This selection is cyclic because it takes place at each cycle before the update of the pseudo-labels, in order to adapt the selected HP to the representation learned by $E$.\\
\paragraph*{One-clustering evaluation} 
We suppose we have access to a separate labeled source validation set $D^\mathcal{S}_{val}$ of $N^\mathcal{S}_{val}$ samples. We also assume that HP search is restricted to a finite size set $\Lambda \subset \mathbb{R}^{n_{\lambda}}$. Given a clustering criterion $\mathcal{L}$ and a HP value $\lambda$ to evaluate, HP tuning phase uses the source empirical risk with samples from $D^\mathcal{S}_{val}$. 
Remember that to satisfy the condition (i) for using the source empirical risk , we should use the whole set of validation samples and on a one-clustering evaluation of the associated risk. Moreover, it can be very computationally expensive to do multiple clusterings to evaluate a unique HP value, and $N^\mathcal{S}_{val}$ can be `too small' to split $D^\mathcal{S}_{val}$ into different subsets for clustering. Therefore, we decide to only perform one clustering on the full set $D^\mathcal{S}_{val}$ to evaluate one parameter value of $\lambda$ with the source empirical risk. At the end of this step, we keep the value $\lambda^*$ that gives the lowest empirical risk value.

\subsection{Learning with conditional domain alignment of feature similarities.}

\subsubsection{Learning features for \reid} 
From the pseudo-labels, the model is trained to minimize a loss function $L_{ID}^{T}$ in order to learn an ID-discriminative feature representation on the target domain. This loss function can be for example the cross entropy loss, the triplet loss, a contrastive loss function or the sum of several of these terms. Besides, we also wish this representation to be ID-discriminative on the source domain by optimizing a loss function $L_{ID}^{S}$ with the labeled source samples. Intuitively, we motivate this choice in order not to degrade the discriminativeness of the representation on the target domain, while optimizing the feature similarity alignment between source and target.

\subsubsection{Domain Discrepancy}
Reducing the domain discrepancy in the conditional similarity feature space is a key aspect to reduce the variance when using the source empirical estimation (as shown in Sec.~\ref{sec:weight_ratio}). Given a differentiable domain alignment criterion $L_{align}$ (e.g., Maximum Mean Discrepancy (MMD) \cite{saito2018maximum}), we optimize the domain alignment in the conditional feature similarity space given by the formula:
 \begin{equation} 
 \label{eq:mmd_loss}
    L^{cond}_{align} =  L_{align}(S^\mathcal{S}_{+},S^\mathcal{T}_{+}) + L_{align}(S^\mathcal{S}_{-},S^\mathcal{T}_{-}) \; ,
\end{equation}
where $S^\mathcal{S}_{+}$, $S^\mathcal{T}_{+}$, $S^\mathcal{S}_{-}$ and $S^\mathcal{T}_{-}$ are the similarities between features of, resp., positive pairs of the source, positive pairs of the target, negative pairs of the source and negative pairs of the target in the feature similarity space. Minimizing this term aligns intra-cluster similarity distributions but also inter-cluster similarity distributions between domains.

\subsubsection{Global criterion} The total loss $L_{total}$ is given by:

\begin{equation}
\label{eq:total_loss}
 L_{total} = L_{ID}^\mathcal{T} +   L_{ID}^\mathcal{S} +   L^{cond}_{align}.
\end{equation}

Note that we choose not to weight the different loss terms in $ L_{total}$ in order not to introduce new additional HP in the UDA context. Indeed, experiments in Sec.~\ref{sec:experiments} will show that this loss choice already allows to get performance improvements from \hypass{} in various UDA benchmarks.

\subsection{General pseudo-code of \hypass{}}

We propose in the Algo.~\ref{algo:general} a pseudo-code for training a pseudo-labeling re-ID UDA framework by using \hypass{}. The proposed automatic hyperparameter tuning from source data (AUTO HP-TUNING) called by Algo.~\ref{algo:general} is detailed in Algo.~\ref{algo:hp} introduced by our approach.\\
Algo.~\ref{algo:general} describes the whole \hypass{} training paradigm. A model is first initialized (INITIALIZATION) to predict the first pseudo-labels for the target training set. Then the algorithm iterates cyclically through a FEATURE EXTRACTION phase with the actual model for the source validation set and the target training set. Then during the AUTO HP-TUNING phase a value for $\lambda^*$ is automatically selected by maximizing a clustering quality criteria. Then this HP value is used to pseudo-label/cluster the target training features during the PSEUDO-LABELING phase. Then the model is fine-tuned with the source training set and the pseudo-labeled target training using \hypass{} loss function (see Eq.~\ref{eq:total_loss}). Algo.~\ref{algo:hp} further details the AUTO HP-TUNING phase, where the algorithm iterates through different HP values proposed by a HP selection strategy or function which are used to pseudo-label the source validation set and compute with the source label a clustering quality metric to be maximized.

\begin{algorithm}[h]
\caption{HyperParameters Automated by Source \& Similarities (\hypass{})}
\begin{algorithmic}
\REQUIRE{Labeled source training set $D^\mathcal{S}$}
\REQUIRE{Labeled source validation set $D^\mathcal{S}_{val}$: $D^\mathcal{S}_{val} \cap D^\mathcal{S} = \varnothing$}  
\REQUIRE{Unlabeled target data $D^\mathcal{T}$} 
\REQUIRE{Clustering/Pseudo-labeling function $C_{\lambda}$ with HP $\lambda$}
\REQUIRE{HP list $\Lambda$}
\REQUIRE{Clustering/Pseudo-Labeling quality metric $\mathcal{L}$ (to maximize)}
\REQUIRE{Loss Functions for Training: $L_{ID}^\mathcal{S}$, $L_{ID}^\mathcal{T}$, $L_{align}$}
\REQUIRE{Number of training epochs $N_{epoch}$}

\REQUIRE{Feature encoder $E$}
\STATE \textbf{INITIALIZATION:} \STATE Compute $S^\mathcal{S}$, $S^\mathcal{T}$  the sets of feature similarities for all pairs of images in  $D^\mathcal{S}$ and $D^\mathcal{T}$, respectively.
\STATE Train $E$ to minimize $L_{init} \gets L_{ID}^\mathcal{S} + L_{align}(S^\mathcal{S},S^\mathcal{T})$.

\STATE \textbf{PSEUDO-LABELING TRAINING:}
\FOR{$t=1$ to $N_{epoch}$}
\STATE \textbf{FEATURE EXTRACTION:} Compute target training features $F^\mathcal{T}$ and source validation features $F^\mathcal{S}_{val}$ from $D^\mathcal{T}$ and $D^\mathcal{S}_{val}$.
\STATE \textbf{AUTO HP-TUNING:} Find $\lambda^*$ that maximizes $\mathcal{L}$ with pseudo-labeling of $F^\mathcal{S}_{val}$  by $C_{\lambda^*}$ and $D^\mathcal{S}_{val}$ ground-truth labels.
\STATE \textbf{PSEUDO-LABELING:} Pseudo-label some/all target samples by $C_{\lambda^*}$ with $F^\mathcal{T}$.
\STATE \textbf{TRAINING:} \STATE Compute $S_{+}^\mathcal{S}$/$S_{-}^\mathcal{S}$, $S_{+}^\mathcal{T}$/$S_{-}^\mathcal{T}$ the positive/negative sets of feature similarities in  $D^\mathcal{S}$ and $D^\mathcal{T}$, respectively.
\STATE Train $E$ to minimize $L_{total} \gets L_{ID}^\mathcal{T} + L_{ID}^\mathcal{S} + L_{align}(S_{+}^\mathcal{S},S_{+}^\mathcal{T}) + L_{align}(S_{-}^\mathcal{S},S_{-}^\mathcal{T})$ with $D^\mathcal{S}$ and pseudo-labeled $D^\mathcal{T}$.
\ENDFOR
\STATE Return $E$
\end{algorithmic}
\label{algo:general}
\end{algorithm}

\begin{algorithm}[h!]
\caption{AUTO HP-TUNING}
\begin{algorithmic}
\REQUIRE{Number of HP values to validate $N_{search}$}
\REQUIRE{Hyperparameter (HP) search function $search\_next()$}
\REQUIRE{Source validation set features $F^\mathcal{S}_{val}$ and labels $Y^\mathcal{S}_{val}$}  
\REQUIRE{Pseudo-labeling function $C_{\lambda^*}$}
\REQUIRE{Pseudo-labeling quality metric $\mathcal{L}$}
\STATE Initialize best HP value $\lambda^*$
\STATE Initialize best metric value $L^* \gets - \infty$
\FOR{$t=1$ to $N_{search}$}
\STATE $\lambda \gets search\_next() $
\STATE Get pseudo-labels $\hat{Y}^\mathcal{S}_{val}$ by clustering $F^\mathcal{S}_{val}$  with $C_{\lambda}$
\STATE Compute $L \gets \mathcal{L}(\hat{Y}^\mathcal{S}_{val},Y^\mathcal{S}_{val})$
\IF{$ L \geq L^*$}
\STATE{$\lambda^* \gets \lambda$}
\STATE{$ L^* \gets L$}
\ENDIF
\ENDFOR
\STATE Return $\lambda^*$
\end{algorithmic}
\label{algo:hp}
\end{algorithm}

\newpage
\section{Experiments}
\label{sec:experiments}
\subsection{Datasets and Protocol}
\label{sec:dataset}

\begin{table}[b!]
\caption{Dataset composition}
\label{table:dataset}
\resizebox{\columnwidth}{!}{
\begin{tabular}{|c|cc|ccc|c|c|}
\hline
Dataset   & \specialcell{train \\IDs} & \specialcell{train \\images} & \specialcell{test\\ IDs} & \specialcell{gallery \\images} & \specialcell{query\\ images} & \specialcell{~ query images \\ per ID} & \specialcell{~ train images \\ per ID} \\ \hline
Market \cite{zheng2015scalable}   & 751       & 12,936        & 750      & 16,364          & 3,368         & 4    & 17   \\
Duke \cite{ristani2016performance}   & 702       & 16,522        & 702      & 16,364          & 2,228         & 3    & 24   \\
PersonX \cite{sun2019dissecting}     & 410       & 9,840        & 856      & 17,661          & 30,816         & 36   & 24    \\
MSMT  \cite{wei2018person}    & 1,041      & 32,621        & 3,060     & 82,161          & 11,659        & 4   & 31   \\ \hline
Vehicle-ID \cite{liu2016deep}  & 13,164     & 113,346       & 800      & 7,332           & 6,532         & 8   & 9    \\
Veri \cite{liu2016deep2}    & 575       & 37,746        & 200      & 49,325          & 1,678         & 8  & 66    \\ 
VehicleX \cite{liu2016deep2}    & 1,362       & 192,150        & N.A.      & N.A.          & N.A.         &  4   &  141   \\ 
\hline
\end{tabular}}
\end{table}

\subsubsection{Datasets} 
We study \hypass{} on different \reid{} adaptation tasks: Person \reid{} and Vehicle \reid.
\textit{Person \reid} is evaluated
on the large \reid{} dataset MSMT17 \cite{wei2018person} (\textit{MSMT}): used as the target domain, it offers a challenging adaptation task due to its large number of images and identities in its gallery (cf. dataset statistics in Tab.~\ref{table:dataset}). We also use Market-1501 \cite{zheng2015scalable} (\textit{Market}) as the target domain using the synthetic dataset PersonX as the source domain. \textit{PersonX} \cite{sun2019dissecting} is composed of synthetic images generated on Unity with different types of person appearances, camera views and occlusions. Then we also report classical benchmarks between Market and DukeMTMC-reID \cite{ristani2016performance} (\textit{Duke}).
\textit{Vehicle \reid{}} is less used than Person \reid{} for UDA \reid{} benchmarking. However, we find it interesting to test our module on a different kind of object of interest and on a potentially different domain discrepancy. We use for this task \textit{Vehicle-ID} \cite{liu2016deep}, Veri-776 \cite{liu2016deep2} (\textit{Veri}) datasets as source or target domains and the synthetic vehicle dataset \textit{VehicleX} \cite{naphade20204th} as source domain.
\subsubsection{Protocol} 
The feature encoder $E$ is evaluated on the target test set. When it is available, we use the source query set as source validation set $D^\mathcal{S}_{val}$ since it is never used elsewhere during the training and no official validation set has been built for these benchmarks. As no test sample is available for VehicleX, we randomly remove 5000 images from the training set to build the validation set. We report the Mean Average Precision (mAP) and rank-1 (top-1) in percents on the target test set after UDA training.\\
\subsubsection{Remarks}
In the different protocols, the source validation sets are very varied in size (number of images) and distinct from the target training set in terms of number of IDs and number of samples per ID. According to our theoretical insights in Sec.~\ref{sec:weight_ratio}, we do not expect these statistic differences to influence a good selection of $\lambda$. 
This will be confirmed by the experiments in Sec.~\ref{sec:validation_exp} for further discussion and experiments about this point and the choice of validation set.
\subsection{Implementation Choices and Details}

\subsubsection{Implementation Choices}
\paragraph{Frameworks} In order to show its effectiveness, we integrate \hypass{} within 3 state-of-the-art methods:
UDAP \cite{song2020unsupervised}, MMT \cite{ge2020mutual} and SpCL \cite{ge2020self}. 
UDAP is a classical pseudo-labeling method, while MMT and SpCL, which manage noise in pseudo-labels, are the best approaches on UDA \reid{}. We focus our experiments on these three frameworks for mainly three reasons: these are renowned re-ID approaches, supplied with a code for reproducibility, and with the best UDA re-ID performance on different adaptation tasks (for SpCL particularly).\\
\paragraph{Clustering algorithm} We focus our experiments on DBSCAN \cite{ester1996density} clustering for two reasons: it is the most widespread in the state of the art and it is used by the best approaches (cf. Sec.~\ref{sec:related}). Thus, our experiments focus on the selection of $\epsilon$ HP that is critical for performance (cf. Sec.~\ref{sec:intro}). 
However, experiments are also made with other clustering algorithms (k-means, Agglomerative Clustering \cite{beeferman2000agglomerative}, HDBSCAN \cite{McInnes2017}) to show the genericity of \hypass{} (cf. Sec.~\ref{sec:ablative}).
The main implementation choices are summarized in Tab.~\ref{table:implementation}. 

\begin{table}[t!]
\caption{Main implementation choices for experiments.}
\label{table:implementation}
\resizebox{\columnwidth}{!}{
\begin{tabular}{c|cl}
Theory                          & Implementation choices                  \\ \cline{1-2}
$\lambda$                       & Maximum Neighborhood Distance $\epsilon$                             \\
$\Lambda$                       & Bayesian Search \cite{gonzalez2016gpyopt} with $\epsilon \in [0,2]   $                    \\
$C_{\lambda}$ & DBSCAN \cite{ester1996density}                                \\
$\mathcal{L}$                   & Adjusted Random Index (ARI) \cite{rand1971objective}                                   \\
$E$                               & ResNet-50 \cite{he2016deep} initialized on ImageNet \cite{deng2009imagenet}    \\
$L_{align}$                     & Maximum Mean Discrepancy (MMD)  \cite{saito2018maximum}       \\
$s$                             & based on $L^2$ distance with normalized features  \\
$L^{S}_{ID}, L^{T}_{ID}$ & \specialcell{Cross-Entropy \& Triplet Losses (UDAP \cite{song2020unsupervised} \& MMT \cite{ge2020mutual})\\ Contrastive Loss (SpCL \cite{ge2020self})}
\end{tabular}
}
\end{table}

\paragraph{Empirical setting comparison}

Pseudo-labeling state-of-the-art approaches use empirical values to set HP $\epsilon$ in DBSCAN. The empirical setting strategy supposes that, in addition to a source labeled dataset, we have access to labels of a part of a calibration target dataset. Therefore, it becomes possible to evaluate the re-ID performance for this cross-dataset adaptation task for different values of $\epsilon$. Then, the $\epsilon$ associated to the best mAP is selected, and reused for other cross-dataset adaptation tasks with another target (unlabeled) dataset.\\
We can choose PersonX as the source dataset. Indeed, PersonX being a synthetic dataset, it is free to label and it does not raise any problem of privacy access to real people identities. For the sake of a robust empirical setting, we suppose that we have access to the test set of MSMT, the biggest and most challenging person re-ID dataset. We train different models with best SOTA method SpCL, for different values of $\epsilon$ ($\epsilon = 0.3, 0.4, 0.5, 0.6, 0.7$ see Fig.~\ref{fig:sensibility}), for the cross-dataset adaptation task PersonX$\rightarrow$MSMT. The mAP of each model is computed on MSMT test set, and the $\epsilon$ associated with the best mAP is kept. After experiments, as shown on Fig.~\ref{fig:sensibility}, we obtain $\epsilon = 0.6$. This value will therefore be reused for other cross-dataset adaptation task, with other target domains, such as
PersonX$\rightarrow$Market.
In Sec.~\ref{sec:ablative}, we compare \hypass{} to this empirical setting strategy (i.e. re-use $\epsilon = 0.6$). Sec.~\ref{sec:effectiveness} gives extensive results for more cross-dataset experiments comparing this empirical strategy with \hypass{}.

\paragraph{HDBSCAN comparison}

HDBSCAN is a hierarchical clustering version of DBSCAN that automatically selects a parameter like $\epsilon$, according to an unsupervised criterion of stability of the clusters in the hierarchy \cite{McInnes2017}. It therefore seems like a reasonable alternative to DSBCAN with empirical setting since it has an unsupervised heuristic to automatically select an $\epsilon$ value. Indeed, we can see HDBSCAN as an automatic HP tuning of $\epsilon$ and it is therefore relevant to compare \hypass{} (DBSCAN) to HDBSCAN on different state-of-the-art methods. The comparison is done in Sec.~\ref{sec:effectiveness}.
HDBSCAN still needs a value for $n_{min}$ controlling the minimum of samples per cluster that is set to 10 during experiments since it gives the best results for different cross-dataset benchmarks in other state-of-the-art work \cite{zhang2019self}.

\subsubsection{Implementation Details}

Our framework is implemented in PyTorch \cite{paszke2019pytorch}. We use 4 x 24Go NVIDIA TITAN RTX GPU for our experiments.\\

\paragraph{Data preprocessing}
\label{sec:preproc} We build two mini-batches: one of size 64 for source images and another of the same size for target ones. Each batch is made of P=16 identities and K=4 instances per identity (and sampled randomly at initialization phase for target due to lack of labels).
Images are resized to 256x128 for person images as in \cite{zheng2015scalable, ristani2016performance, wei2018person}  and 224x224 for vehicle ones as in \cite{liu2016deep,liu2016deep2}.  We randomly flipped and cropped images but we do not use random erasing augmentation during initialization phase since it has been shown to be harmful for direct transfer \cite{luo2019bag}. \\

\paragraph{Feature encoder/Network} For state-of-the-art comparison, we use a Resnet-50 \cite{he2016deep} pretrained on ImageNet \cite{deng2009imagenet} as our backbone. The last stride of ResNet-50 is set to 2 to have higher resolution feature map. After the global average pooling layer, we add a BatchNorm layer and then the classification layer(s) which is initialized with the Kaiming initialization \cite{he2016deep}. At test time, we use the normalized 2048 pre-classification features with squared Euclidean distance to compute the ranking lists.\\

\paragraph{Domain Alignment} For $L_{align}$, we use the MMD PyTorch implementation of D-MMD paper \cite{mekhazni2020unsupervised} with the Gaussian kernel  \footnote{https://github.com/djidje/D-MMD}. The features are normalized before computing the (conditional) pairwise feature similarities.\\

\paragraph{Initial phase} The network is trained during 60 epochs. The learning rate is set to $3,5.10^{-4}$ and is decayed by a factor 10 every 20 epochs. Since we have not yet pseudo-labels for the target data, the classical Cross Entropy Loss and Triplet Loss are optimized on the source samples only, jointly with $L_{align}$ on the source and target unlabaled samples.\\

\paragraph{HP tuning} We perform HP search with Bayesian optimization. We choose Bayesian optimization since it is a powerful HP search approach that is able to look for relevant HP values ($\Lambda$) according to an updated belief \cite{gonzalez2016gpyopt}. We use the library GPyOpt \footnote{https://sheffieldml.github.io/GPyOpt/} using Gaussian processes. We just used the default Bayesian optimizer parameters using basic Gaussian processes as the modeling function and Expected Improvement (EI) as the acquisition type. The search range for $\epsilon$ is set to $[0,2]$ (it is the whole range of variation for $\epsilon$ since the features are normalized and thus belong to the unit hypersphere). For k-means variant, $k$ is searched in the full range [1, number of target training samples]. At each Auto HP tuning step, we evaluate $N_{HP} = 50$ hyperparameter values proposed by the Bayesian search. With this setting, the initial value can be sampled randomly since it has no influence on performance as shown later in Sec.~\ref{sec:variants}.\\
The Adjusted Random Index (ARI) \cite{hubert1985comparing} is computed between the source validation set ground truth labels and the cluster predictions using the scikit-learn implementation \footnote{https://scikit-learn.org/}.\\

\paragraph{Pseudo-labeling training phase} Implementation details for this step are framework-specific. We put the symbol "*" after the name of the framework to indicate that it corresponds to our version (to include \hypass{} and allow easier experimental comparisons) based on the original framework. We give the specific implementation details below. If not specified we make the same choices (optimizer, number of epochs,...) as given in their respective paper.

\subsubsection{Framework-specific details}

\paragraph{UDAP*} We build our code from the UDAP \cite{song2020unsupervised} implementation publicly available on the official UDAP GitHub \footnote{https://github.com/LcDog/DomainAdaptiveReID}. For UDAP, we use an initialization phase before the pseudo-labeling UDA learning. DBSCAN is run on k-reciprocal encoded features with $k=30$ whereas the k-means version directly uses the feature as in the original paper. The minimum samples $n_{min}$ per cluster is set to 4 (as in paper \cite{song2020unsupervised}). Compared to the UDAP paper, we use only one 2048 feature space with Triplet Loss, and add a Cross Entropy Classification loss for the target pseudo-labeled samples (since it improves performance).
To add \hypass{}, we add to this UDAP* loss, the classification  and triplet losses $L_{ID}^\mathcal{S}$ for the source samples (by initializing a new classification layer for source IDs) as well as $L_{align}$. Other training hyparameters are the same as in the UDAP paper \cite{song2020unsupervised}.
 
\paragraph{MMT*} We build our code from the MMT \cite{ge2020mutual} implementation publicly available on the official MMT GitHub \footnote{https://github.com/yxgeee/MMT}. For MMT, we use an initialization phase before the pseudo-labeling UDA learning. DBSCAN is run on k-reciprocal encoded features with $k=30$ whereas the k-means version directly uses the features as in the original paper \cite{song2020unsupervised}. The minimum samples $n_{min}$ per cluster is set to 4. To add \hypass{}, we only add to the original MMT global loss function, the hard classification  and triplet losses defined in paper \cite{song2020unsupervised}, for the source samples (by initializing a new classification layer for source IDs), as well as $L_{align}$. Other training hyparameters are the same as in MMT paper \cite{song2020unsupervised}.
 
\paragraph{SpCL*} We build our code from the SpCL \cite{ge2020self} implementation publicly available on the official SpCL GitHub \footnote{https://github.com/yxgeee/SpCL}. It does not need an initialization phase and the ID loss on source samples is already implemented and used in the original framework with the contrastive loss. To include \hypass{}, we add $L_{align}$ to the global objective and remove the cluster criterion (for \hypass{} and HDBSCAN experiments). Other hyparameters are the same as in the SpCL paper \cite{ge2020self}.
\\
\newline
Our implementations based on the authors' code for UDAP*, MMT* gives better performance than those reported in the papers. For SpCL*, we obtained only slightly inferior performance (-1.1 p.p. at worst), which should not interfere with conclusions that will be made from experiments in Sec.~\ref{sec:exp_analysis}.

\begin{center}
\begin{table*}[t!]
\caption{Comparison of \hypass{} with empirical setting strategy on pseudo-labeling state-of-the-art methods on person \reid{} adaptation tasks. * means  we used authors’ code and add \hypass{}.
}
\label{table:state-of-the-art}
\resizebox{\linewidth}{!}{
\begin{tabular}{c|c|cc|cc|cc|cc|cc|}
\cline{2-12}
\multirow{2}{*}{} &
  \multirow{2}{*}{HP selection} &
  \multicolumn{2}{c|}{Market$\rightarrow$MSMT} &
  \multicolumn{2}{c|}{PersonX$\rightarrow$Market} &
  \multicolumn{2}{c|}{PersonX$\rightarrow$MSMT} &
  \multicolumn{2}{c|}{Market$\rightarrow$Duke} &
  \multicolumn{2}{c|}{Duke$\rightarrow$Market} \\ \cline{3-12} 
                                             &                            & mAP  & rank-1 & mAP  & rank-1 & mAP  & rank-1 & mAP  & rank-1 & mAP & rank-1 \\ \hline
\multicolumn{1}{|c|}{\multirow{2}{*}{UDAP* \cite{song2020unsupervised}}} & Empirical ($\epsilon=0.6$) & 12.0 & 30.6     & 48.4 & 68.4     & 10.5 & 26.3 & 50.1 & 70.2 & 55.3 & 78.1    \\ 
\multicolumn{1}{|c|}{} & HDBSCAN & 11.8 & 29.8     & 48.1 & 68.3     & 10.3 & 25.9 & 51.3 & 72.5 & 55.9 & 80.0    \\ \cline{2-12} 
\multicolumn{1}{|c|}{} &
  \textbf{\hypass{}} &
  \textbf{21.4} &
  \textbf{48.8} &
  \textbf{62.2} &
  \textbf{73.7} &
  \textbf{15.6} &
  \textbf{36.4} & \textbf{64.9} & \textbf{78.0} & \textbf{69.8} & \textbf{87.1} \\ \hline \hline
\multicolumn{1}{|c|}{\multirow{2}{*}{MMT* \cite{ge2020mutual}}} &
  Empirical ($\epsilon=0.6$) &
  23.8 &
  49.9 &
  71.1 &
  66.8 &
  17.4 &
  39.0 & 65.3 & 78.1 & 73.6 & 89.4 \\
\multicolumn{1}{|c|}{} &
  HDBSCAN &
  23.0 &
  47.8 &
  70.9 &
  66.1 &
  18.0 &
  41.1 & 65.2 & 78.2 & 74.2 & 90.4 \\ \cline{2-12}
\multicolumn{1}{|c|}{} &
  \textbf{\hypass{}} &
  \textbf{25.1} &
  \textbf{52.2} &
  \textbf{74.5} &
  \textbf{88.9} &
  \textbf{20.3} &
  \textbf{45.9} & \textbf{68.8} & \textbf{82.8} & \textbf{76.0} & \textbf{90.1} \\ \hline \hline
\multicolumn{1}{|c|}{\multirow{3}{*}{SpCL* \cite{ge2020self}}} & Empirical ($\epsilon=0.6$) & 25.7  & 53.4  & 72.2 & 86.1   & 22.1 & 47.7      & 68.3 & 82.5 & 76.1 & 89.8 \\
\multicolumn{1}{|c|}{}                       & HDBSCAN & 24.6  & 52.0  & 70.8 & 86.5   & 21.1 & 46.9      & 66.4 & 81.3 & 75.8 & 89.5 \\ \cline{2-12} 
\multicolumn{1}{|c|}{}                       & \textbf{\hypass{}}                    & \textbf{27.4} & \textbf{55.0}     & \textbf{77.9} & \textbf{91.5}     & \textbf{23.7} & \textbf{48.6}     & \textbf{71.1} & \textbf{84.5} & \textbf{78.9} & \textbf{92.1}\\    \hline

\end{tabular}
}
\end{table*}
\end{center}

\begin{center}
\begin{table}[t!]
\caption{Comparison of \hypass{} with empirical setting strategy on pseudo-labeling state-of-the-art methods on vehicle \reid{} adaptation tasks. * means  we used authors’ code and add \hypass{}.
}
\label{table:state-of-the-art_v}
\resizebox{\columnwidth}{!}{
\begin{tabular}{c|c|cc|cc|}
\cline{2-6}
\multirow{2}{*}{} &
  \multirow{2}{*}{HP selection} &
  \multicolumn{2}{c|}{VehicleID$\rightarrow$Veri} &
  \multicolumn{2}{c|}{VehicleX$\rightarrow$Veri} \\ \cline{3-6} 
                                             &                            & mAP & rank-1 & mAP & rank-1 \\ \hline
\multicolumn{1}{|c|}{\multirow{2}{*}{UDAP* \cite{song2020unsupervised}}} &
  Empirical ($\epsilon=0.6$) & 35.6 & 74.1     & 35.0  & 75.9     \\
\multicolumn{1}{|c|}{} &
  HDBSCAN & 35.9 & 75.0     & 35.5  & 79.9     \\ \cline{2-6} 
\multicolumn{1}{|c|}{} &
  \textbf{\hypass{}} &
  \textbf{36.9} &
  \textbf{74.9} &
  \textbf{37.0} &
  \textbf{77.0} \\ \hline \hline
\multicolumn{1}{|c|}{\multirow{2}{*}{MMT* \cite{ge2020mutual}}} &
  Empirical ($\epsilon=0.6$) &
  36.4 &
  74.2 &
  36.3 &
  75.8 \\
\multicolumn{1}{|c|}{} &
 HDBSCAN &
  37.0 &
  75.9 &
  36.5 &
  75.9 \\ \cline{2-6} 
\multicolumn{1}{|c|}{} &
  \textbf{\hypass{}} &
  \textbf{36.9} &
  \textbf{75.0} &
  \textbf{36.8} &
  \textbf{76.1} \\ \hline \hline
\multicolumn{1}{|c|}{\multirow{3}{*}{SpCL* \cite{ge2020self}}} & Empirical ($\epsilon=0.6$) & 37.6 & 79.7     & 37.4 & 81.0         \\
\multicolumn{1}{|c|}{}                       & HDBSCAN & 37.4 & 79.9     & 37.5 & 79.8        \\ \cline{2-6} 
\multicolumn{1}{|c|}{}                       & \textbf{\hypass{}}  &            \textbf{40.0}   & \textbf{81.1}     & \textbf{40.3} & \textbf{81.9}\\    \hline

\end{tabular}
}
\end{table}
\end{center}
\newpage
\section{Results and analysis of \hypass{}.}
\label{sec:exp_analysis}

\subsection{Effectiveness of \hypass{} on state-of-the-art methods.}
\label{sec:effectiveness}

\subsubsection{Performance analysis of \hypass{}.}
\paragraph{HyPASS vs empirical setting.}
Results in resp.
Tab.~\ref{table:state-of-the-art} and Tab.~\ref{table:state-of-the-art_v} show that our automatic HP selection improves the three SOTA frameworks, on all person \reid{} and vehicle \reid{} adaptation tasks. This improvement is particularly significant for UDAP: it increases, e.g., the mAP by +9.4 p.p.
on Market$\rightarrow$MSMT and +13.8 p.p. on PersonX$\rightarrow$Market over the empirical setting strategy. This improvement of using \hypass{} over the empirical setting strategy is also consistent for "easier" adaptation tasks such as Duke$\rightarrow$Market (+14.5 p.p.) and Market$\rightarrow$Duke (+14.8 p.p.).   \hypass{} seems thus to benefit a simple pseudo-labeling approach like UDAP by making it competitive with more complex 
approaches like MMT, designed to be resistant to pseudo-label noise.
Our contribution also improves consistently MMT and SpCL (the best state-of-the-art approaches) on all tasks: there is, e.g., up to +4.1 p.p. mAP improvement on PersonX$\rightarrow$Market for SpCL compared to the SpCL reported performance (using empirical setting). Furthermore, we highlight that SpCL with \hypass{} for cross-dataset UDA re-ID is able to outperform (or at least be competitive with) performance of the latest UDA re-ID and unsupervised approaches: for example, SpCL + \hypass{} reaches 71.1 \% mAP on Market$\rightarrow$Duke whereas \cite{Xuan_2021_CVPR, Yang_2021_CVPR,Zhang_2021_CVPR,Zheng_2021_CVPR, Chen_2021_CVPR, zheng2021exploiting} reach respectively, 59.1\%, 53.8\%, 69.2\%, 69.2\%, 69.1\% and 67.6\% mAP on Duke or Market$\rightarrow$Duke.\\
We recall that experiments have been conducted with an empirical setting performed on PersonX$\rightarrow$MSMT ($\epsilon = 0.6$). A different empirical setting choice, on PersonX$\rightarrow$Market for example, would let to an empirical value $\epsilon = 0.5$ (see Fig.~\ref{fig:sensibility}), and therefore improvements given by using HyPASS would be greater on other cross-datasets. Indeed, with $\epsilon = 0.5$, the performance are further degraded for SpCL on PersonX$\rightarrow$MSMT (20.3\% mAP). Therefore \hypass{} improves the mAP by +3.4 p.p with this other empirical setting for SpCL.

\paragraph{HyPASS vs HDBSCAN.}
Moreover, results in Tab.~\ref{table:state-of-the-art} and Tab.~\ref{table:state-of-the-art_v} show that using \hypass{} (with DBSCAN) consistently outperforms HDBSCAN for the three frameworks and on all the person \& vehicle re-ID cross-datasets benchmarks. Indeed, results show that HDBSCAN is in fact not necessarily better than using the empirical setting $\epsilon = 0.6$ (for e.g. 24.6\% mAP for SpCL on PersonX$\rightarrow$Market with HDBSCAN instead of 25.7\% mAP with empirical setting) or only brings small improvements (+0.1 p.p. for SpCL and PersonX$\rightarrow$Market with HDBSCAN instead of empirical setting). Therefore, the conclusions done for empircal setting vs HyPASS remains the same for emprical setting vs HDBSCAN: among those three HP selection strategies, using \hypass{} appears to be the best one. 

\subsection{A cluster quality analysis to understand the effectiveness of HyPASS.}

\label{sec:cluster_quality}
\begin{figure}[b!]
\includegraphics[width=\columnwidth]{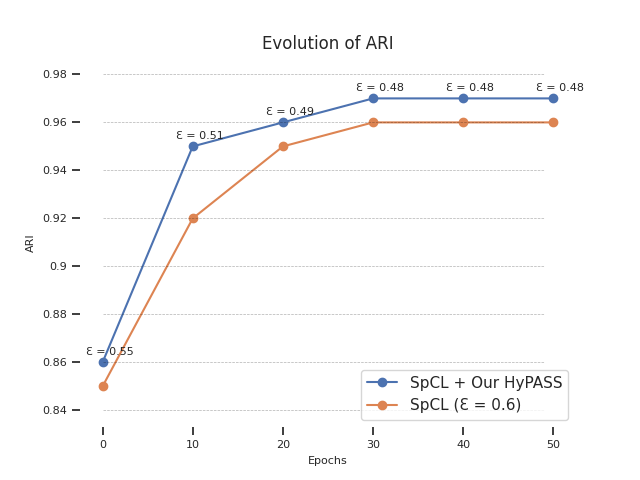}
\caption{Positive impact of an iterative HP tuning of $\epsilon$ (HyPASS) on the clustering quality. The figure represents evolution of ARI of the pseudo-labeled target training set through epochs on PersonX$\rightarrow$Market with SpCL \cite{ge2020self}.}
\label{fig:ari}
\end{figure}
To understand more precisely the positive impact of \hypass{} on the training process,
we monitor the evolution of:
(i) the quality of the clusters found during the pseudo-labeling cycles, through the ARI of the pseudo-labeled target samples, every 10 epochs (after the pseudo-labels are updated);
(ii) HP $\epsilon$ found by \hypass{}.
Fig.~\ref{fig:ari} shows that \hypass{} seems to find better clusters (with better ARI) than the fixed empirical parameter strategy ($\epsilon = 0.6$) from the first epochs on. We believe this impact on the quality of the clusters is `iterative': better clusters (pseudo-labels) in early epochs will imply the learning of better representations and therefore the possibility to make better clusters when the pseudo-labels are updated. Fig.~\ref{fig:ari} also highlights that the value of the selected $\epsilon$ changes cyclically (as the feature representation changes) over the pseudo-labeling cycles.\\

\subsection{Ablative Study \& Parameter Analysis on training time and performance.} 

\subsubsection{Relevance of the optimization losses} 
\label{sec:ablative}
In the ablative study presented in Tab.~\ref{table:study}, we seek to verify the relevance of our optimization losses (see Eq.~\ref{eq:total_loss}) for the selection of HP for the UDAP~\cite{song2020unsupervised}, MMT~\cite{ge2020mutual} and SpCL~\cite{ge2020self} approaches.

We train
different variants
by removing terms from the total loss function (see. Eq.~\ref{eq:total_loss}) in order to observe their effects on the final performance (mAP). Variant \#5 corresponds to \hypass{} with the total loss function.
First, we notice that training the model to be discriminating on the source domain (variant \#3) together with our Auto HP tuning allows improvements compared to variant \#1 (only Auto HP tuning) for UDAP:
+18.6 p.p. mAP on PersonX$\rightarrow$Market. We believe that the feature encoder in variant \#1 specializes on target domain while forgetting source domain initialization. Thus, HP selection becomes worse because it is done on a representation that is less and less discriminating  for the source domain over time.
After a certain number of epochs, bad choices of HP may impact the quality of pseudo-labels and, then, target representation. In variant \#2, performance drops even more if alignment is added without $L_{ID}^{S}$ (variant \#2): -28.7 p.p. on PersonX$\rightarrow$Market
. We believe that alignment on poorly discriminative source is even more harmful to the target representation. We notice the same behavior for MMT with -29.4 p.p. and -8.7 p.p. respectively.
Therefore, when using Auto HP of \hypass{}, it is necessary to keep optimizing source ID-discriminative features with $L_{ID}^{S}$.

Adding the term $L^{cond}_{align}$ of conditional domain alignment of feature similarities (variant \#5) further improves substantially performance by using Auto HP (variant \#3): +13.4 p.p. on PersonX$\rightarrow$Market
. The same improvement trend is observed for MMT and SpCL. This seems to confirm our theoretical considerations of reducing the variance of the estimation by reducing the domain discrepancy in the feature similarity space when using Auto HP (see Sec.~\ref{sec:feature_sim}).

Finally, by comparing variants \#4 and \#5, we observe the contribution of our cyclic Auto HP: +9 p.p. on PersonX$\rightarrow$Market
. The same is true for MMT and SpCL. We believe this shows the importance of choosing a suitable HP for each pseudo-labeling update cycle as done with the Auto HP tuning step of \hypass{} (variant \#5).

\begin{table}[t]
\caption{Ablation studies on \hypass{} for UDAP*, MMT* and SpCL* methods (mAP in \%). \#5 is (full) \hypass{}.}
\label{table:study}
\resizebox{\columnwidth}{!}{
\begin{tabular}{c|c|ccc|c|c|}
\cline{2-7}
\multirow{2}{*}{} & \multirow{2}{*}{{ \#}}& \multicolumn{3}{c|}{Losses}              & \multirow{2}{*}{\begin{tabular}{c}\small{Auto.}\\ \small HP tuning \end{tabular} } & \multicolumn{1}{c|}{\begin{tabular}{c} PersonX\\$\rightarrow$Market \end{tabular}}
\\  \cline{7-7}
\multirow{2}{*}{} & &$L_{ID}^{T}$ & $L_{ID}^{S}$ & $L^{cond}_{align}$ &   & mAP  
\\ \cline{1-7}
\multicolumn{1}{|c|}{\multirow{5}{*}{UDAP* \cite{song2020unsupervised}}} & 1 & \checkmark       &        &       &      \checkmark      & 30,2           
\\
\multicolumn{1}{|c|}{} &\multicolumn{1}{|c|}{2} & \checkmark        &          & \checkmark      & \checkmark         & 20.1            
\\
\multicolumn{1}{|c|}{}                       &\multicolumn{1}{|c|}{3} & \checkmark        & \checkmark         &       & \checkmark         & 48.8
\\
\multicolumn{1}{|c|}{}                       &\multicolumn{1}{|c|}{4} & \checkmark        & \checkmark        &   \checkmark      &          & 53.2  
\\
\multicolumn{1}{|c|}{}                       &\multicolumn{1}{|c|}{5} & \checkmark        &  \checkmark          &   \checkmark       &  \checkmark            & \textbf{62.2}                    
\\ \cline{1-7}
\multicolumn{1}{|c|}{\multirow{5}{*}{MMT* \cite{ge2020mutual}}} & 1 & \checkmark       &        &       &      \checkmark      & 55.9          
\\
\multicolumn{1}{|c|}{}                       &\multicolumn{1}{|c|}{2} & \checkmark        &          & \checkmark      & \checkmark        & 41.3 
\\
\multicolumn{1}{|c|}{}                       &\multicolumn{1}{|c|}{3} & \checkmark        & \checkmark         &       & \checkmark         & 70.7
\\
\multicolumn{1}{|c|}{}                       &\multicolumn{1}{|c|}{4} & \checkmark        & \checkmark        &   \checkmark      &          & 71.5          
\\
\multicolumn{1}{|c|}{}                       &\multicolumn{1}{|c|}{5} & \checkmark        &  \checkmark          &   \checkmark       &  \checkmark            & \textbf{74.5}           
\\ \cline{1-7}
\multicolumn{1}{|c|}{\multirow{3}{*}{SpCL* \cite{ge2020self}}} & 3 & \checkmark        & \checkmark         &       & \checkmark         & 68.1
\\
\multicolumn{1}{|c|}{}                       &\multicolumn{1}{|c|}{4} & \checkmark        & \checkmark        &   \checkmark      &          & 73.9            
\\
\multicolumn{1}{|c|}{}                       &\multicolumn{1}{|c|}{5} & \checkmark        &  \checkmark          &   \checkmark       &  \checkmark            & \textbf{77.9}                     
\\ \cline{1-7}

\end{tabular}}
\end{table}

\subsubsection{Performance of \hypass{} with other clustering algorithms.}
\label{sec:clustering}

\begin{figure}[t]
\includegraphics[width=1\columnwidth]{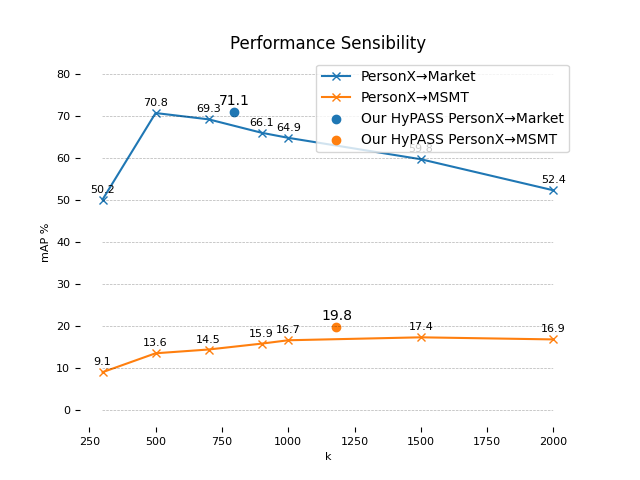}
\caption{Performance sensibility for the state-of-the-art framework MMT \cite{ge2020mutual} with respect to $k$ parameter of k-means.}
    \label{fig:sensibility_mmt}
\end{figure}

\paragraph{K-means} 
Other clustering algorithms can be used instead of DBSCAN. But they still need to set HP. For example, k-means relies on the number $k$ of clusters. Similarly to the sensibility of DBSCAN with $\epsilon$, Fig.~\ref{fig:sensibility_mmt} shows that the performance with k-means is also sensible to the number of clusters HP. Again, choosing a good HP value is crucial to get good performance: for example, by choosing k = 250, performance drops from 70.8\% to 50.2\% mAP for PersonX$\rightarrow$Market and from 16.6\% to 10.1\% compared to k = 500 for PersonX$\rightarrow$MSMT with MMT
.\\
Therefore, empirical setting strategy for choosing k on another adaptation task quite limits performance too. Indeed, re-using the best value from PersonX$\rightarrow$MSMT (k=1500) leads to 59.8\% mAP for PersonX$\rightarrow$Market whereas it could have been 70.8\% for k=500. Reciprocally, choosing k=500 from PersonX$\rightarrow$Market leads to 13.6\% for PersonX$\rightarrow$MSMT instead of 17.4\% for k=1500.\\
As illustrated on Fig.~\ref{fig:sensibility_mmt} and shown in Tab.~\ref{table:cluster}, using \hypass{} leads to better performance compared to empirical setting. For PersonX$\rightarrow$Market with MMT, it leads to 71.1\% mAP instead of 59.8\% reusing k=1500 obtained by empirical setting on PersonX$\rightarrow$MSMT.
\paragraph{Agglomerative Clustering} Agglomerative Clustering \cite{beeferman2000agglomerative} is another clustering algorithm that can be used instead of DBSCAN. As DBSCAN, Agglomerative Clustering is a density-based clustering algorithm that relies on a neighborhood distance threshold parameter $\epsilon$. \hypass{} can also improve performance of pseudo-labeling methods using this clustering algorithm. As shown in Tab.~\ref{table:cluster}, for PersonX$\rightarrow$Market with SpCL, using \hypass{} leads to 78.2\% mAP instead of 72.2\% using empirical setting ($\epsilon = 0.6$).

\begin{table}[t!]
\caption{Performance (mAP) of \hypass{} on k-means and Agglomerative Clustering and with state-of-the-art pseudo-labeling approaches. We set $k=1500$ as empirical setting since it is the best configuration on PersonX$\rightarrow$MSMT in our experiments. For Agglomerative Clustering, empirical setting $\epsilon = 0.6$ is motivated by analogy to our experiments for PersonX$\rightarrow$MSMT with DBSCAN (see Fig.~\ref{fig:sensibility}) which is also a density-based algorithm.}
\label{table:cluster}
\resizebox{\columnwidth}{!}
{
\begin{tabular}{c|c|c|c|}
\cline{2-4}
                       & Clustering & HP choice       & PersonX$\rightarrow$Market \\ \hline
\multicolumn{1}{|c|}{\multirow{2}{*}{MMT* \cite{ge2020mutual}}}  & \multirow{2}{*}{k-means}           & Empirical $k=1500$         & 59.8 \\ \cline{3-4}
\multicolumn{1}{|c|}{} &            & \textbf{\hypass{}} & \textbf{71.1}              \\ \hline \hline
\multicolumn{1}{|c|}{\multirow{2}{*}{SpCL* \cite{ge2020self}}} & \multirow{2}{*}{Agglo. Clustering \cite{beeferman2000agglomerative}} & Empirical $\epsilon = 0.6$ & 72.6 \\ \cline{3-4} 
\multicolumn{1}{|c|}{} &            & \textbf{\hypass{}} & \textbf{78.2}              \\ \hline
\end{tabular}}
\end{table}

\subsubsection{Influence of the validation set size.}
\label{sec:validation_exp}

\begin{table}[h]
\caption{Experiments with different validation set size $N^S_{val}$  on SpCL for PersonX$\rightarrow$Market showing the validation set size on performance and training computation time.}
\label{table:time}
\resizebox{\columnwidth}{!}
{
\begin{tabular}{c|c|cccc|}
\cline{2-6} 
& \multicolumn{1}{|c|}{\multirow{2}{*}{Empirical setting}} & \multicolumn{4}{c|}{HyPASS $N^S_{val} =$} \\
           &  & 1000  & 5000  & 10000 & 30816 \\ \cline{1-6}
\multicolumn{1}{|c|}{Time}        & 60h12 (6 $\times \sim$ 10h02)    & 12h08 & 34h39 & 42h21 & 68h43 \\ \cline{1-6}
\multicolumn{1}{|c|}{mAP (in \%)} & 72.2       & 76.1  & 77.8  & 77.8  & 77.9  \\ \hline
\end{tabular}
}
\end{table}

We have seen that \hypass{} brings consistent improvements on various adaptation tasks (see Tab.~\ref{table:state-of-the-art}) and therefore with various sizes validation set (see Tab.~\ref{table:dataset} number of query validation images). These also show experimentally that performance improvements from \hypass{} is also robust to various dataset compositional bias between the source and target domains, more particularly the difference in number of query per IDs and IDs.\\
But the validation set size also intuitively influences the clustering computation time, and thus the full training computation time of the frameworks where \hypass{} is added. Moreover, it is also interesting to have more experimental insights on the influence of the validation set on performance improvement of \hypass{} for a fixed adaptation task. That's why we further investigate the influence of the validation set size on the training computation time and the \reid{} performance. Experiments are conducted on PersonX$\rightarrow$Market for the SpCL framework. For this, we randomly select $N$ images from PersonX query set. The execution time (on the same machine) and \reid{} performance are reported in Tab.~\ref{table:time}. The empirical setting strategy has been performed on PersonX$\rightarrow$Market adaptation task with the 5 HP values: $\epsilon = 0.3, 0.4, 0.5, 0.6, 0.7$. The empirical setting strategy requires 5 training of SpCL with the 5 HP values for PersonX$\rightarrow$MSMT then one more training of SpCL for PersonX$\rightarrow$Market with the best $\epsilon$ ($\epsilon$ = 0.6) evaluated by the mAP on the target test set.  \\
We notice that the training computation time increases with the validation set size. However, it is still fairly reasonable for a training time including hyperparameter selection. Even with a large validation set (30k images), the complete training time lasts only 68h40 and brings significant performance for this adaptation task (+5.7 p.p.). In practice, it is quite big for a validation set size, and experiments show that even with 5k images, performance remains the same, with a training computation time reduced by about 25h33 compared to empirical setting. More generally, performance of \hypass{} is not really sensible validation set size variations tested (from 1/10 up to 3 times the size of the training set induced only 1.8 p.p. variation). Indeed, this is consistent with our guess that reducing the domain discrepancy should allow less sensibility to the number of validation samples, as motivated by Eq.~\ref{eq:variance} in Sec.~\ref{sec:variance}.

\subsubsection{Influence of Auto  HP selection criterion.}
\label{sec:time}

We included in the design of \hypass{} different modeling choices aiming at improving training time and performance. To show the relevance of these choices, we conducted various experiments by changing \hypass{} HP selection strategy on PersonX$\rightarrow$MSMT on the framework SpCL.
First, \hypass{} HP selection is directly based on cyclic clustering quality evaluations instead of \reid{} performance evaluation in order to reduce the computation cost. As illustrated in Tab.~\ref{table:comput}, using \hypass{} but with the mAP criterion (the \reid{} criterion which is our main task) on the source test to select the clustering HP gives almost the same performance of \hypass{} (77.1\% mAP), but  greatly increases the training time to 90h29 instead of 68h43. We reckon that it is mainly due to the higher number of training steps needed to evaluate HP values with the mAP. Even though the best target mAP is the final goal, our assumption to select HP by clustering quality evaluation instead of mAP evaluation (Sec.~\ref{sec:cyclic}) is relevant to limit the training time while having the best \reid{} performance.

\begin{table}[h]
\caption{Impact of \hypass{} with different version of Auto HP criterion on the \reid{} performance and computation. Experiments done on SpCL for PersonX$\rightarrow$Market.}
\label{table:comput}
\resizebox{\columnwidth}{!}
{
\begin{tabular}{|c|c|c|c|}
\hline
Variants                 & Auto HP criterion & mAP  & Time  \\ \hline
SpCL w/ mAP HP selection & re-ID task (mAP)  & 77.1 & 90h29 \\ \hline
SpCL w/ HyPASS           & clustering task (ARI)  & 77.9 & 68h43 \\ \hline
\end{tabular}
}
\end{table}

\subsubsection{Performance with other \hypass{} variants}
\label{sec:variants}

\paragraph{Domain Alignment.}
We conducted some experiments with other implementation choices for \hypass{} with SpCL on PersonX$\rightarrow$Market. For instance, a 2-layer Domain Adversarial Neural Network (DANN \cite{ganin2016domain}) can be used instead of MMD to align the pairwise feature similarities. Tab.~\ref{table:variants} shows that \hypass{} keeps performance improvement over the framework without \hypass{} (+5.1 p.p. mAP compared to SpCL without \hypass{}). 
\paragraph{Cluster quality criterion.}
The Normalized Mutual Information (NMI) can replace the ARI and gives as good performance (+0.2 p.p. mAP in Tab.~\ref{table:variants} compared to \hypass{} with ARI). 
\paragraph{HP search strategy.}
Using a more simple HP search strategy like grid search for $\epsilon \in \Lambda = [0.05,0.1,0.15,...,2]$ can replace the Bayesian search. It still gives good results with \hypass{} (-0.3 p.p. compared to Bayesian search in Tab.~\ref{table:variants}).
\paragraph{HP search initialization.}
In Tab.~\ref{table:init}, when using Bayesian Search with $N_{HP} = 50$ proposed values per HP tuning phase, the initial value $\epsilon_0$ has completely no impact on performance.

\begin{table}[h]
\caption{Performance of \hypass{} for PersonX$\rightarrow$Market with SpCL* \cite{ge2020self} pseudo-labeling method on different variants.}
\label{table:variants}
\centering
\resizebox{\columnwidth}{!}
{
\begin{tabular}{c|c|}
\cline{2-2}
                                                & PersonX$\rightarrow$Market (mAP in \%)\\ \hline
\multicolumn{1}{|c|}{SpCL*}               & 72.2           \\ \hline
\multicolumn{1}{|c|}{SpCL*+ \hypass{} (MMD + Bay. search + ARI)}               & 77.9           \\ \hline
\multicolumn{1}{|c|}{SpCL* + \hypass{} (DANN \cite{ganin2016domain} + Bay. search + ARI)}               & 77.3           \\ \hline
\multicolumn{1}{|c|}{SpCL* + \hypass{} (MMD + Bay. search + NMI)}         & 78.1           \\ \hline
\multicolumn{1}{|c|}{SpCL* + \hypass{} (MMD + Grid Search + ARI)} & 77.6           \\ \hline
\end{tabular}}
\end{table}

\begin{table}[h]
\caption{Robustness of \hypass{} against the Bayesian search initialization of $\epsilon_{0}$. Performance for PersonX$\rightarrow$Market of \hypass{} with SpCL* \cite{ge2020self} pseudo-labeling method are reported with different values of Bayesian Search initialization.}
\label{table:init}
\centering
\resizebox{ \columnwidth}{!}
{
\begin{tabular}{|c|c|}
\cline{1-2}   Bayesian search initialization $\epsilon_{0}$ &PersonX$\rightarrow$Market (mAP in \%) \\ \hline
\multicolumn{1}{|c|}{0.01} & 77.8           \\ \hline
\multicolumn{1}{|c|}{0.8}  & 77.9           \\ \hline
\multicolumn{1}{|c|}{2}    & 77.8           \\ \hline
\end{tabular}}
\end{table}

\section{Conclusion}
\label{sec:conclusion}
This paper addresses the problem of empirical HP selection for pseudo-labeling UDA re-ID approaches as it can have a negative impact on performance when addressing new unlabeled cross-datasets. We provided novel theoretical insights to highlight the conditions under which a source-based selection is effective for the UDA clustering task. These allowed us to design a new method, \hypass{}, to automatically select suitable HP for the clustering phase of pseudo-labeling UDA methods. It is based on source guidance and domain similarity alignment. When \hypass{} is applied to select critical clustering HP instead of using empirical settings, it consistently improves performance of 
the best methods of the state-of-the-art.
We believe that suitable HP selection could be relevant for the unsupervised re-ID scenario in which pseudo-labeling methods seem also effective \cite{ge2020self}. Further work could be done on how to select suitable HP in the unsupervised scenario when we don't want to use any available labeled dataset.

\newpage

\begin{IEEEbiography}[{\includegraphics[width=1in,height=1.25in,clip,keepaspectratio]{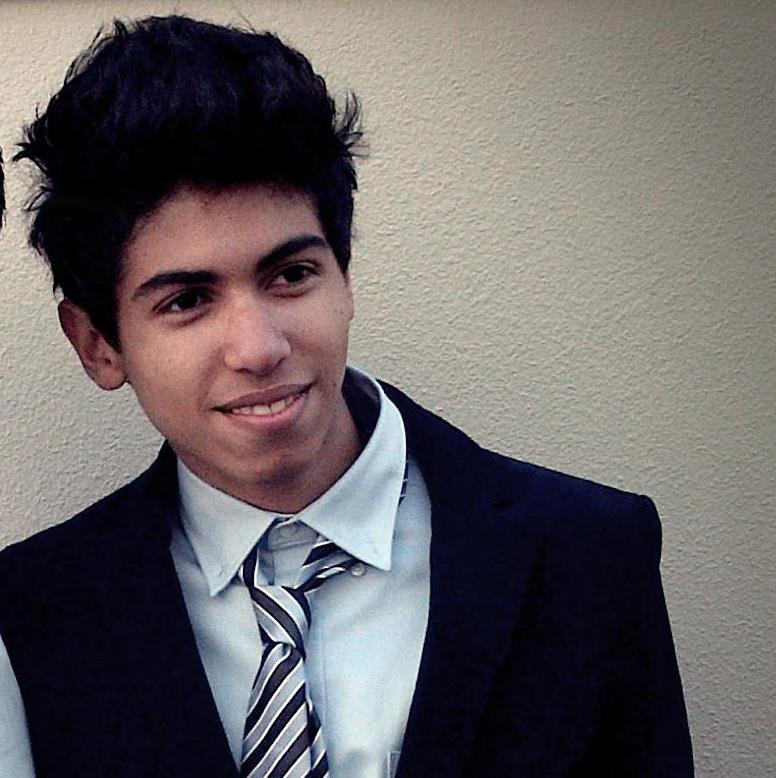}}]{Fabian Dubourvieux}
Fabian Dubourvieux is actually a PhD student in computer vision and machine learning at CEA LIST, Paris-Saclay University and the LITIS CNRS Laboratory, INSA Rouen Normandie. His main research interests include unsupervised domain adaptation, self-supervised learning and object re-identification.
\end{IEEEbiography}

\begin{IEEEbiography}[{\includegraphics[width=1in,height=1.25in,clip,keepaspectratio]{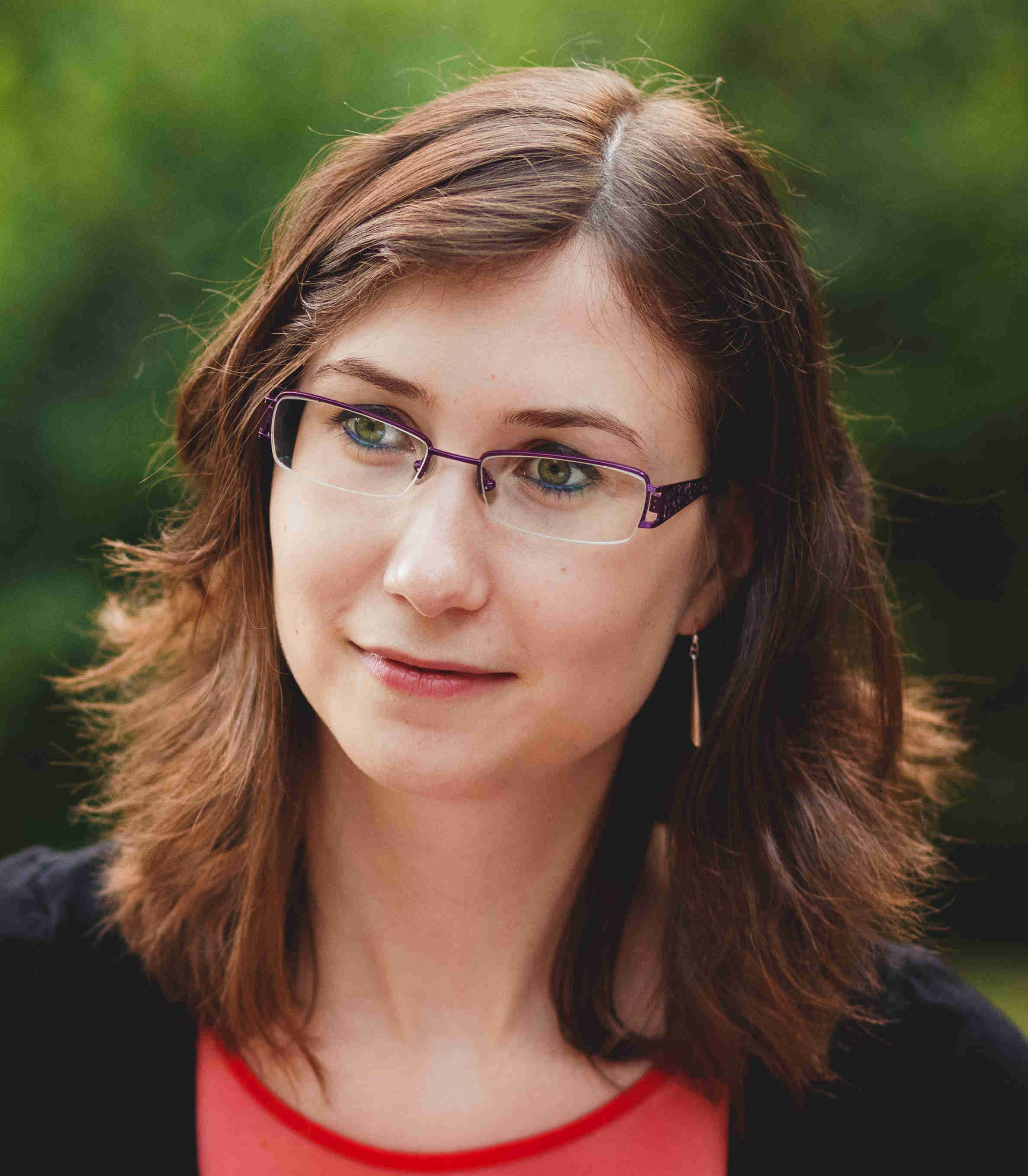}}]{Angélique Loesch}
Angélique Loesch received PhD degree in computer science from the University Blaise Pascal, Clermont-Ferrand, France in 2017. She did her research about Simultaneous Localization and Mapping (SLAM) and 3D object tracking in partnership with the Pascal Institute. She is currently a permanent researcher at CEA LIST, Paris-Saclay University. Her main research interests include computer vision with a focus on visual perception with deep learning (object re-identification, instance and  semantic segmentation, few-shot classification and detection).
\end{IEEEbiography}

\printbibliography
\begin{IEEEbiography}[{\includegraphics[width=1in,height=1.25in,clip,keepaspectratio]{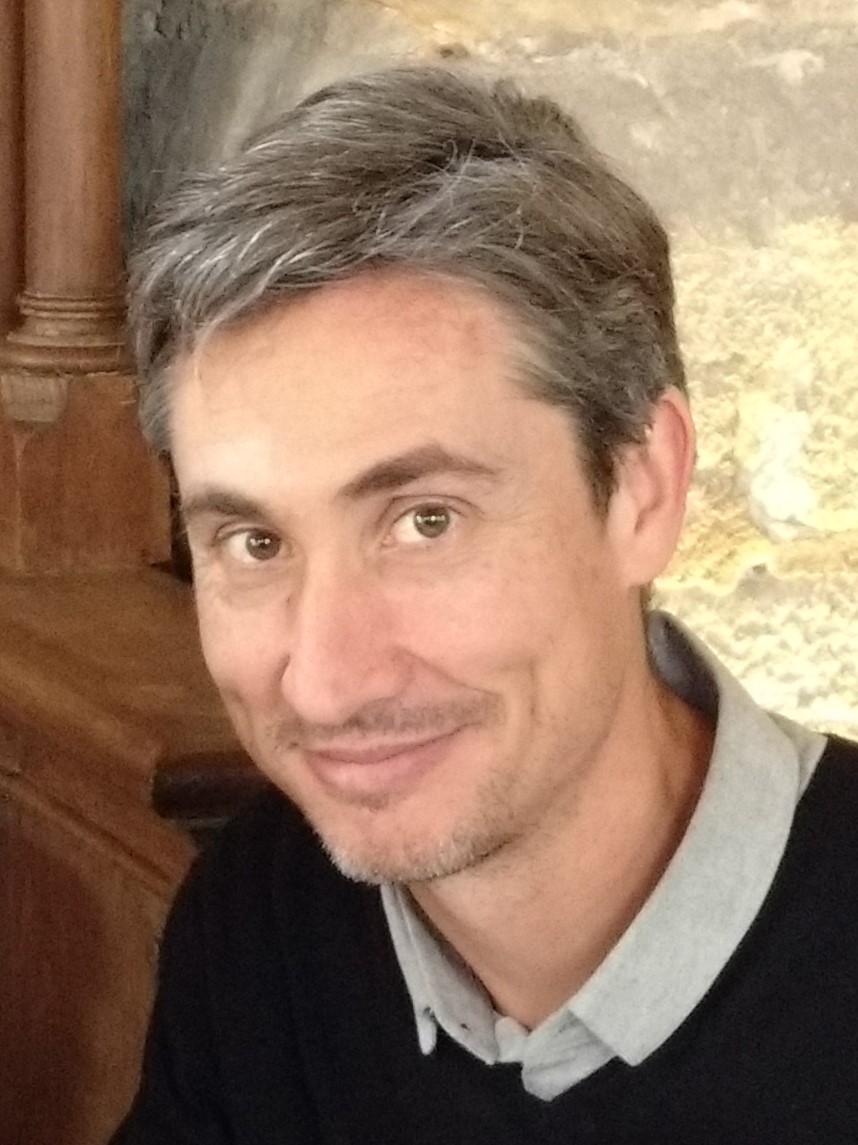}}]{Romaric Audigier}
Romaric Audigier is a researcher in computer vision \& machine learning at CEA LIST, Paris-Saclay University. He received a PhD in image processing from the State University of Campinas (UNICAMP) in 2007. After a post-doctoral position at Mines-ParisTech, he joined CEA LIST in 2009. His current research interests include frugal learning paradigms like unsupervised domain adaptation and few-shot learning applied to visual scene analysis tasks (object detection, segmentation, re-identification, tracking, human interaction and event detection).
\end{IEEEbiography}

\begin{IEEEbiography}[{\includegraphics[width=1in,height=1.25in,clip,keepaspectratio]{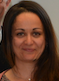}}]{Samia Ainouz}
Samia Ainouz is a full Professor at INSA Rouen Normandy. She is the head of the team Intelligent Transportation systems since June 2019. Her research area is around Multimodality for intelligent vehicle navigation including data fusion, 3D reconstruction, VSLAM. She supervised 7 PhD students around road scene analysis and autonomous navigation. Recently, she focused her research towards non-conventional imaging for autonomous navigation in adverse weather conditions using Deep learning tools. She is currently the head of the ANR project ICUB dealing with road scene analysis in adverse weather conditions with collaboration with Peugeot PSA, Stereolabs and ImVia.
\end{IEEEbiography}

\begin{IEEEbiography}[{\includegraphics[width=1in,height=1.25in,clip,keepaspectratio]{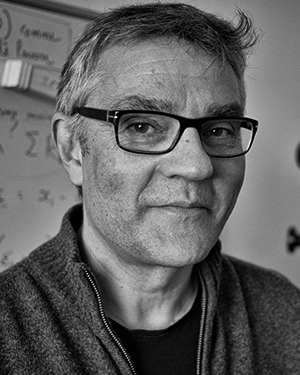}}]{Stéphane Canu}
Stéphane Canu is a Professor of the LITIS research laboratory and of the information technology department, at the National institute of applied science in Rouen (INSA).
 He has been the dean of the computer engineering department he create in 1998 until 2002 when he was named director of the computing service and facilities unit. In 2004 he join for one sabbatical year the machine learning group at ANU/NICTA (Canberra) with Alex Smola and Bob Williamson.  In the last five years, he has published approximately thirty papers in refereed  conference proceedings or journals in the areas of theory, algorithms and applications  using kernel machines learning algorithm and other flexible regression methods.  His research interests includes deep learning, kernels machines, regularization, machine learning applied to signal processing, pattern classification, factorization for recommender systems and learning for context aware applications.
\end{IEEEbiography}

\EOD

\end{document}